\documentclass{files/IEEEtran}
\IEEEoverridecommandlockouts

\usepackage{cite}
\usepackage{amsmath,amssymb,amsfonts}
\usepackage{algorithmic}
\usepackage{graphicx}
\usepackage{textcomp}
\usepackage{xcolor}
\usepackage[acronym]{glossaries} 
\usepackage[binary-units]{siunitx} 
\usepackage{booktabs} 
\usepackage[shortlabels, inline]{enumitem} 
\usepackage{multicol} 
\usepackage{subcaption}

\usepackage{files/customCommands} 
\usepackage{files/paperCommands} 

\usepackage[hidelinks]{hyperref}

\def\thesubsubsectiondis{\unskip\arabic{subsubsection})}

\def\BibTeX{{\rm B\kern-.05em{\sc i\kern-.025em b}\kern-.08em
    T\kern-.1667em\lower.7ex\hbox{E}\kern-.125emX}}

\begin{document}

\title{Borinot: an open 
thrust-torque-controlled robot\\ 
for 
research on
agile aerial-contact motion}

\author{Josep Mart\'i-Saumell \quad Hugo Duarte \quad Patrick Grosch  \\ Juan Andrade-Cetto \quad Angel Santamaria-Navarro \quad Joan Sol\`a 
\thanks{All authors are with the Institut de Rob\`otica i Inform\`atica Industrial, CSIC-UPC, Llorens Artigas 4-6, Barcelona 08028 (e-mail: \{jmarti, \}@iri.upc.edu). This work was partially supported by the Spanish Ministry of Science and Innovation under the projects EBCON (PID2020-119244GB-I00, funded by MCIN/ AEI /10.13039/501100011033); and AUDEL (TED2021-131759A-I00, funded by MCIN/ AEI /10.13039/501100011033 and by the "European Union NextGenerationEU/PRTR"); and by the Consolidated Research Group RAIG (2021 SGR 00510) of the Departament de Recerca i Universitats de la Generalitat de Catalunya.}
}
\maketitle


\begin{abstract}

This paper introduces Borinot, an open-source aerial robotic platform designed to conduct research on hybrid agile locomotion and manipulation using flight and contacts. 
This platform features an agile and powerful hexarotor that can be outfitted with torque-actuated limbs of diverse architecture, allowing for whole-body dynamic control. 
As a result, Borinot can perform agile tasks such as aggressive or acrobatic maneuvers with the participation of the whole-body dynamics.

The limbs attached to Borinot can be utilized in various ways; 
during contact, they can be used as legs to create contact-based locomotion, or as arms to manipulate objects. 
In free flight, they can be used as tails to contribute to dynamics, mimicking the movements of many animals. 
This allows for any hybridization of these dynamic modes, making Borinot an ideal open-source platform for research on hybrid aerial-contact agile motion.

To demonstrate the key capabilities of Borinot in terms of agility with hybrid motion modes, we have fitted a planar 2DoF limb and implemented a whole-body torque-level model-predictive-control. 
The result is a capable and adaptable platform that, we believe, opens up new avenues of research in the field of agile robotics. Interesting links\footnote{Documentation: \url{www.iri.upc.edu/borinot}}\footnote{Video: \url{https://youtu.be/Ob7IIVB6P_A}}.
\end{abstract}


\section{Introduction}
\label{sec:introduction}

The field of robotics is currently witnessing a notable surge in interest regarding the understanding and development of intricate and dynamic motion~\cite{Rubenson_SR2022, Ajanic_SR2020, foehn_AgiliciousOpensourceOpenhardware_2022}. 
This exploration is not only fascinating from a biomechanical perspective but also vital for effectively constructing and controlling inherently unstable robots, such as humanoids or aerial robots. 
This is particularly evident in the realm of  legged robotics (humanoids and quadrupeds), where all the multi-articulated body parts interact dynamically to generate overall motion, and consequently, locomotion and manipulation tasks are interdependent, necessitating the consideration of the concept of loco-manipulation and the existence of hybrid motion modes. 
In aerial robotics, the emergence of \glspl{uam} has increased the complexity of aerial robots with multiple articulations~\cite{Ollero_2022}. 
This presents new opportunities for hybrid motion, such as aerial loco-manipulation utilizing contacts, the action of tail inertia to modify flight dynamics (as observed in studies like \cite{Nabeshima-2019-arque-tail,schwaner-2021-tail-reorient,tang-2022-quadruped-tail}), or mixed locomotion modes like the jump-and-fly technique exhibited by certain animals such as chickens and locusts \cite{wei-19-fly-jump,badri-2022-birdbot,birn2014don,tobalske2007aerodynamics}. 
However, the topic of agile aerial loco-manipulation using contacts remains largely unexplored, and  research on \glspl{uam} usually concentrates only on the manipulation aspects, which are tackled at relatively low dynamics.

\begin{figure}[t]
\centering
\includegraphics[width=0.7\columnwidth]{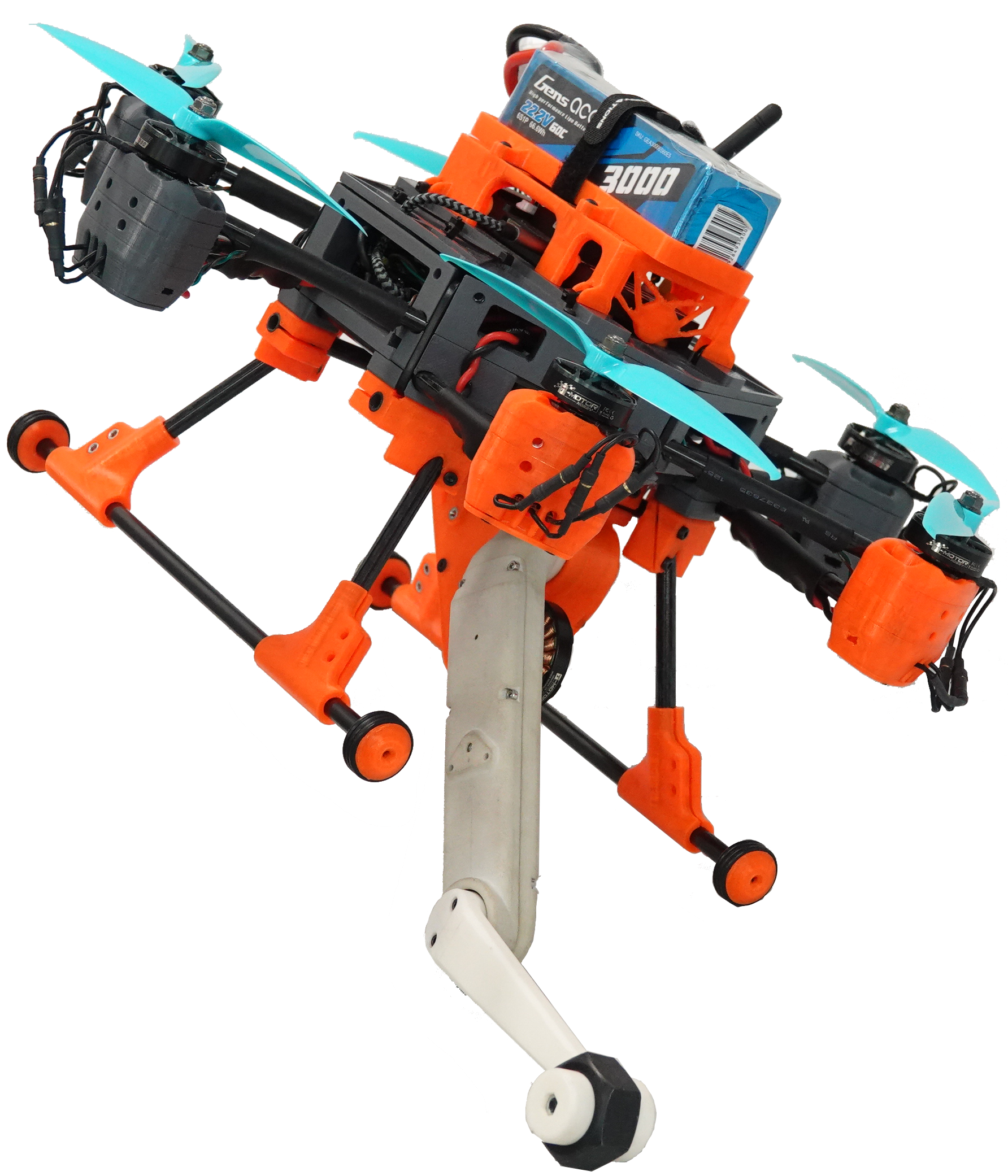}
\caption{The open-source  platform Borinot for agile whole-body torque-controlled flying and contact loco-manipulation.}
\label{fig:borinot}
\end{figure}

In view of this, the main objective of this paper is to lay the groundwork for the study of whole-body agility at the convergence of aerial and legged robotics. 
We begin by carefully examining the essential characteristics that an agile aerial loco-manipulator robot should possess. 
To support our definition, we construct a practical prototype named Borinot (\figRef{fig:borinot}), which we make openly available through a comprehensive open-source initiative. 
Furthermore, we demonstrate the robot's capabilities by performing various typical movements that illustrate the use of its limb as a tail, as an arm, and as a leg, by means of a relatively straightforward \gls{mpc} controller. 
This investigation opens up a broad research area where we will subsequently share our progress in more advanced controllers, such as better MPC designs or reinforcement learning (RL).

This paper is organized as follows.
In the rest of this section, we explore the related work.
In  \secref{sec:conditions-agile-robots}, we elaborate on the conditions for building agile aerial loco-manipulators. 
In Secs. \ref{sec:hardware} and \ref{sec:software} we describe the  hardware and software of Borinot. 
In \secref{sec:control_architecture}, we present a preliminary design of a \gls{mpc} controller, which allows us to validate Borinot in terms of a capable agile platform.
In \secref{sec:experiments}, we present hardware tests relevant to an agile motion and real executions of different agile loco-manipulation tasks. 
Sec. \ref{sec:discussion} closes with a discussion.

\subsection{Related work}
\label{sec:related-work}

Morphologically speaking, the kind of robot that most resembles our proposed robot is a \gls{uam}, a flying platform endowed with one or more robotic arms. 
So far, agility has not been seriously explored in the field of aerial manipulation and there is no \gls{uam} allowing for agile movements \cite{ruggiero_2018, Ollero_2022}.
Current \glspl{uam} equip arms with high gear ratios, hence not compliant, which are actuated using low-torque motors.
This prevents these robots to be controlled dynamically. 
Initially, \gls{uam}s were controlled in a decentralized manner, meaning that platform and arm controls are independent \cite{ruggiero_multilayer_2015,santamaria_nmpc_2017,rossi_trajgeneration_2017,lipiello_visualservoing_2016}.
This requires decoupled dynamics of platform and arm, resulting in quasi-static operations. 
Centralized whole-body control has been typically implemented using feedback linearization \cite{rajappa_tilthex_2015,tognon_control-aware_2018,kim_stabilizing_2018}, sometimes exploiting the differential flatness properties of the system  \cite{yuksel_differential_2016,tognon_dynamic_2017}. 
The drawback of feedback linearization is that it works at the kinematic level, and it is not obvious how to include the natural dynamics of multi-articulated bodies.

In many cases, \gls{uam}s have been built by adding arms to  existing commercial \glspl{uav}.
This helps solving a significant amount of practical problems related to flight, both in hardware and control, especially those regarding safety via recovery flight modes.
However, the capabilities of \gls{uav}s meant for commercial purposes are not always aligned with those needed for research and it is difficult to find manufacturers. 
For example, the dynamic control of such platforms demands direct control of individual rotor velocities, which is not always available.
In addition, in most cases these flying platforms have been withdrawn due to a reduced market (\eg~Asctec, and DracoR from Uvify).
This lack and/or inconvenience of platforms has led the robotics laboratories to create their own, although not many of them are open-source.
A recently published open-source \gls{uav} is the Agilicious drone~\cite{foehn_AgiliciousOpensourceOpenhardware_2022}, which is designed for agile motion but has no manipulation capabilities.
Non-commercial UAMs have been described in~\cite{Saeed_2018, Paul_2018, Perez-Jimenez-2020, Suarez_2020}, although they mount traditional arms with high gear ratios and still require quasi-static maneuvering.
Other existing works focus on presenting unconventional flying platform designs such as~\cite{Nguyen_2018, Park_2018} but they are not open-source.
In summary, new hardware proposals are required to grant \glspl{uam} with agile capabilities, for both platforms and arms, using also new control paradigms enabling the exploitation of the whole-body dynamics.

In many relevant aspects, it is more clarifying to regard Borinot not as an evolution of the \gls{uam}s, but rather of the legged robots, which are now allowed to fly.
Indeed, research on legged robots (quadrupeds, bipeds and humanoids) with strong dynamic capabilities is very mature, and already contemplates most of the requirements for agility \cite{hutter-16-anymal,feng-14-atlas,mastalli-20-crocoddyl,Carpentier_2018,grimminger_OpenTorqueControlledModular_2020} (see \secref{sec:conditions-agile-robots} for these requirements). 
In particular, and in contrast with what we said about \gls{uam}s, many legged robots are already conceived with compliant torque-controlled limbs, and their control schemes have been contemplating the whole-body dynamics already for a while.
It is therefore very reasonable, and this is precisely what we do, to draw from the state-of-the-art in contact locomotion research \cite{Carpentier_2018,mastalli-20-crocoddyl} and adapt its techniques to agile aerial loco-manipulation. 
Hardware-wise, the high dynamics nature of the actuators used in legged robots make them a natural choice for agile aerial maneuvering.
For example, Borinot takes advantage of the open dynamic robot initiative (ODRI\footnote{https://open-dynamic-robot-initiative.github.io/}) and its actuator design~\cite{grimminger_OpenTorqueControlledModular_2020} to conceive different kinds of powerful, light, compliant and torque-controlled limbs. 
As for the agile base platform, we have to design and build one from the ground up.

\subsection{Beyond the state-of-the-art}
Endowing legged robots with flying capabilities opens up a new area of research in the confluence of legged robotics and aerial manipulation. 
Unlike \glspl{uam}, such robots are now not meant exclusively for manipulation.
Interestingly, they can be used for research on agile aerial manipulation \textit{and} locomotion, and all the possible hybrid motion modes that may emerge from this new viewpoint.
Put otherwise, we do not concentrate on the semantics of the intended tasks (\eg~walk, jump, grasp, pick-and-place), but on the nature of the motion itself (the exploitation and control of the dynamic interactions between objects with mass).
One well-known robot intended for hybrid motion is Leonardo \cite{Kim_2021}, a bipedal hybrid robot that can walk and fly. 
The limitation of Leonardo is that it is built without agility or dynamism in mind, again resorting to limbs with high gear ratios, thereby suffering from the same limitations of the \glspl{uam} described above. 
With the introduction of Borinot we aim to overcome all these limitations.

\section{The agile aerial loco-manipulator}
\label{sec:conditions-agile-robots}

Borinot is designed with agility in mind. 
For this reason, we believe it is necessary first to introduce our notion of agility, providing a detailed definition from which we can work on.
To do so, we start with the dictionary definition and refine it so that we can engineer it. 
We then discuss how such a concept of agility can be implemented in robots, both in terms of electro-mechanics and control, to finally conclude with a relation of the characteristics that we believe are required for conceiving agile aerial loco-manipulators.

\subsection{The concept of agility}

The concept of agility is defined in the dictionary\footnote{Oxford learner’s dictionaries: agility, accessed on April 3,
2023 at \url{https://www.oxfordlearnersdictionaries.com/definition/english/agility}} as  \textit{``the ability to move quickly and easily''}.
This definition is considerably vague as it overlooks some crucial factors that are key for robotics. 
The existence of a complex multi-articulated body with torso, head, and limbs is taken for granted in natural language and biomechanical studies when referring to agility, and so it is implicit in the definition, but requires careful attention when engineering machines. 
Two characteristics of such bodies that must be considered are under-actuation and redundancy. 
Under-actuation means the system cannot be commanded to follow arbitrary trajectories in configuration space, requiring the generation of non-trivial maneuvers~\cite{underactuated}. 
Redundancy means there are multiple ways to satisfy a task, and thus, certain \glspl{dof} remain available for secondary tasks.

A third characteristic of these bodies that demands consideration is that they often rely on contacts with the environment to generate all or a significant part of the forces that will produce the motion.

With all these considerations made explicit, we shall enrich the definition of agility as \textit{``the ability of a complex body to combine quick maneuvers easily, using contacts and other forces against the environment''}. 
This explicit definition outlines the major necessary conditions for designing and controlling an agile robot, with each keyword representing a specific characteristic. 
For instance, the term \textit{`maneuver'} demands strong prediction capabilities, while \textit{`quick'} emphasizes dynamism. 
\textit{`Easy'} necessitates the minimization of effort, hence the importance of optimality when controlling the robot. 
Executing quick maneuvers with minimal effort requires the utilization of the \textit{`complex body'}'s natural dynamics: we must carefully predict those forces that will modify the robot's whole-body dynamics as little as necessary while accomplishing the desired tasks. 
These forces have to be produced by the robot, hence the need for force-torque actuation and control.
Whole-body control is also demanded by the keyword \textit{`combine'}, in the sense that the robot can do several things at once, such as flipping over while looking at a specific target. 
\textit{`Contacts}`, and especially unexpected or poorly predicted contacts in the context of high dynamics, demand light and compliant limbs to minimize the undesired effects of energetic impacts, but powerful enough to be used for locomotion.
Finally, to ensure easy dynamics, robots with high \gls{com} and small moment of inertia are preferred, with actuators placed where their effect is maximized.

\subsection{Degrees of agile motion}
\label{sec:degrees-agility}

To provide effective design guidelines, we must also consider different degrees of agility. 
In this regard, we can subdivide agile motion into categories and examine how they impact the requirements of the desired robot.
First, `soft' or `gentle' motion is characterized by small actuation effort and results in the execution of tasks with low dynamic content. 
It is the consequence of minimizing energy.
Second, `graceful' motion would employ a significant part of the control range but avoids frequent saturation and harsh control changes. 
It is a trade-off between energy and time.
Finally, `acrobatic' or even `aggressive' motion would be where more powerful actuators are often saturated, allowing sharp transitions between saturation extremes. 
It comes from the minimization of time. 
The upper limit for aggressiveness results in switched, permanently saturated actuation, which can be dealt with time-optimal control strategies, yielding actuation of the type bang-bang \cite{romero-2021-time-optimal-aerial}.
Clearly, a more powerful robot can be driven more aggressively, and therefore the overall ratio between motor power and mass (or inertia) will determine the final character of the motion a robot will be able to achieve.

\subsection{Types of agile aerial-contact loco-manipulation}
\label{subsec:types_agile_locoman}

We now discuss the concepts of locomotion and manipulation and their impact on the design and control of aerial robots. 

When there are no contacts involved, \textit{locomotion} refers to plain flight, with the limb functioning as a tail, while \textit{manipulation} involves modifying non-dynamic characteristics such as aesthetics or information (for tasks like spraying, filming, or spying), using the limb as an arm.

In the presence of contacts, various situations arise depending on the ratio of inertias between the robot and the object being contacted:

    - Locomotion occurs when the contact is utilized to alter the robot's motion, such as walking or jumping. 
    It is assumed that the inertia of the contacted object is infinitely larger than that of the robot. 
    In this case, the limb functions as a leg.

    - A first manipulation class involves using contact to influence the object's motion. 
    This corresponds to traditional object manipulation tasks like grasping, pick-and-place, pushing, or pulling. 
    Ideally, it is assumed that the object's inertia is negligibly small compared to the robot's. 
    Here, the limb acts as an arm.

    - A second manipulation class employs contact to modify non-dynamic characteristics, such as drawing or measuring. 
    In this scenario, the object's inertia is considered infinitely larger than the robot's, resulting in the interaction forces affecting the robot's motion.

We observe that it is the ratio of inertia between the robot and the object that determines whether the contact will affect the motion of the robot, the object, or both --- this stems directly from the principle of conservation of momentum. 
When the inertias are comparable, any interaction between the robot and the object will influence both dynamics, resulting in a hybrid motion combining locomotion and manipulation elements. 
This hybridization is particularly interesting in highly dynamic scenarios. 
We can refer to this as hybrid loco-manipulation, such as when a robot bounces off or intercepts an airborne object.
Similarly, we can consider hybrid motion modes such as jump-and-flight locomotion, which, as we show at the end of this paper, achieves much higher efficiency than plain flight.

Conceiving a robot for such situations and controlling it requires a unified approach encompassing loco-manipulation and considering dynamism as its core aspect. 
This approach emphasizes the nature of the motion and interactions themselves, focusing on how to apply forces to achieve the desired dynamic effects on the robot \textit{and/or} the object, rather than the specific categorization (locomotion versus manipulation) of the intended action.

\subsection{Required and desired characteristics}
\label{subsec:int_requirements}

Given all the ideas presented so far, we summarize a set of required and desired characteristics for conceiving agile aerial loco-manipulator robots, particularly when they are devoted to research.

\subsubsection{\bf{\Gls{mc} platform}}
\begin{enumerate}[(a)]
  \item \emph{Small sized, compact}: Provides smaller inertia. Allows operation in smaller indoor research arenas.
  \item \emph{High \gls{com}}: allows easier tilting for more agile maneuvers.
  \item \emph{High \gls{twr}}: 
  The \gls{twr} is used in \cite{foehn_AgiliciousOpensourceOpenhardware_2022} to quantify the agility of a platform. 
  \item \emph{Direct motor control}: Control has to act directly on forces, and so we need to command propeller thrust. A direct command over motor velocity or similar is also practicable as long as a good mapping onto thrust is available.
\end{enumerate}

\subsubsection{\bf{Robotic limbs}}
\begin{enumerate}[a)]
  \item \emph{Lightweight}: To undermine as less as possible the overall \gls{twr} of the assembled loco-manipulator.
  \item \emph{Compliant, low friction, low gear ratios}: To avoid the introduction of non-conservative forces, such as friction. Also, to minimize the disrupting effects of impacts.
  \item \emph{High-torque motors}: To allow aggressive maneuvers. To support (a significant amount of) the robot's weight for locomotion. Important since we have low gear ratios.
  \item \emph{Torque-controllable}: 
  Torque can be either measured with a torque sensor at the joint, which is expensive in terms of weight and price, or estimated from the motor current. 
  The last option is only possible with the low friction provided by low gear ratios.
\end{enumerate}

\subsubsection{\bf{Control}} 
\begin{enumerate}[a)]
    \item \emph{Predictive}: To be able to generate maneuvers. 
    \item \emph{Optimal}: To be able to move easily. Optimality can be model-based (such as \gls{mpc}) or data-based (evolutionary algorithms, learning).
    \item \emph{Whole-body}: To account for all dynamic interactions between the parts.
    \item \emph{Torque-based}: To act on the dynamics of the robot.
    \item \emph{Phase switching}: To account for the appearance and disappearance of contacts.
\end{enumerate}

\subsubsection{\bf{Other desired characteristics}}

\begin{enumerate}[a)]
  \item \emph{Open-source}: Researchers should have access to the hardware and firmware designs. 
  Open-source platforms may have large communities of developers who can provide support and contribute to the platform's growth.
  \item \emph{Easy to build, cost-effective, flexible}: The  Hardware should be 3D-printable and require common tools for assembly. 
  This reduces costs related to production, maintenance and repairs, and allows easy customization.
  \item \emph{Conceptually simple}: Avoid design complications such as non-planar \glspl{mc} or coaxial propellers. They move the focus of research away from agility.
\end{enumerate}

\section{Borinot's hardware}
\label{sec:hardware}

\subsection{Design overview}

Borinot (bumblebee in Catalan, \figRef{fig:borinot}) is conceived as a compact and powerful thrust-controlled hexacopter to which different light and compliant torque-driven limbs \cite{grimminger_OpenTorqueControlledModular_2020} can be outfitted. 
The \gls{com} is located high to allow rapid and easy destabilization from the hovering stance. 
It has a powerful control unit constituted by an Intel NUC i7, able to perform high-demanding control tasks such as whole-body force-torque-based \gls{mpc} involving contacts.

Borinot comprises two bodies (\figRef{fig:ro_platform_overview}): The upper body is a hexacopter flying platform. The lower body comprises one or more robotic limbs and the landing gear.
The following gives a detailed overview of both subsystems and how they are assembled.

\subsection{Upper body: hexacopter platform}

\begin{figure*}[t]
  \centering
  \includegraphics[width=0.25\linewidth]{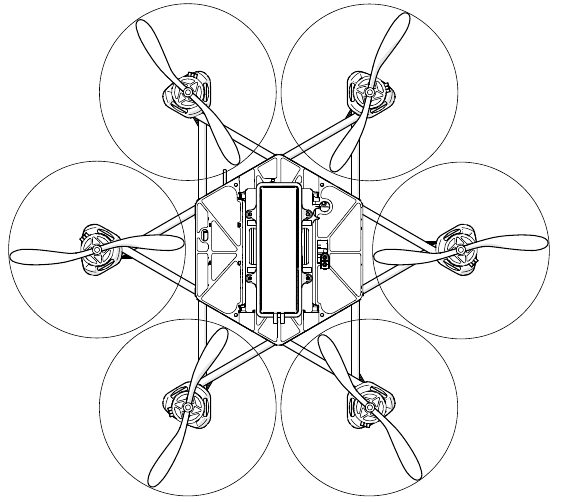}
  \includegraphics[width=0.39\linewidth]{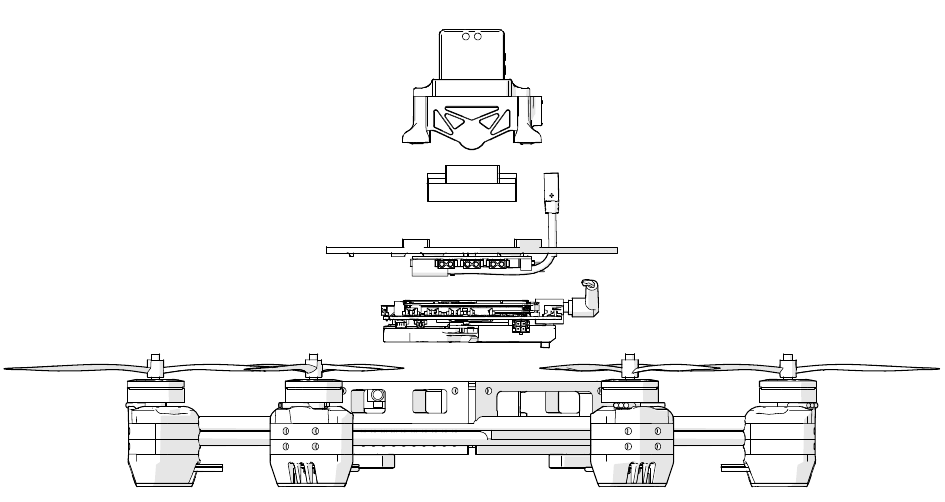}
  \includegraphics[width=0.32\linewidth]{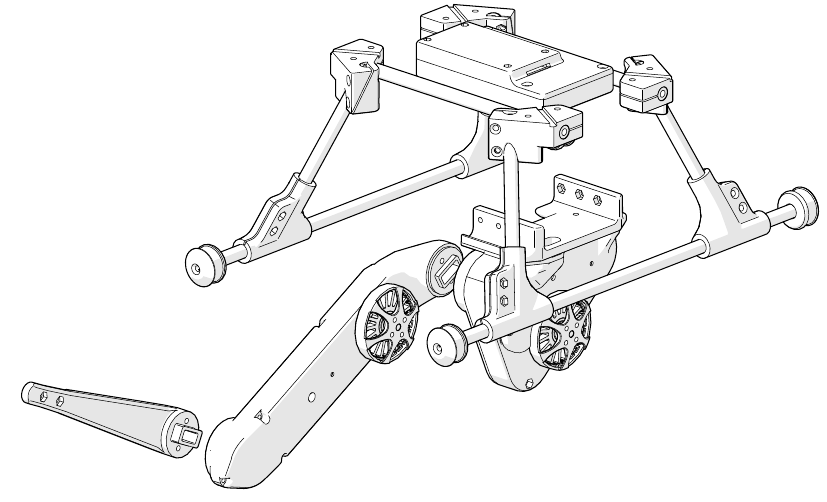}
  \caption{Borinot loco-manipulator hardware. 
  \emph{Left:} Top-view of the base, showing the carbon tubes hexagram star, the six propellers, and the central hexagon. 
  \emph{Center:} Exploded front view of the base, showing from bottom to top, frame with motor pods, NUC, cover with the power module, PixHawk controller, and battery. 
  \emph{Right:} landing gear with articulated 2 \gls{dof} limb (acting as arm, leg, or tail).}
  \label{fig:ro_platform_overview}
\end{figure*}

The Borinot base (\figref{fig:ro_platform_overview}, left and center) has been designed as a hexacopter.
This maximizes compactness, allowing its use in small-sized robotics laboratories.
The structural frame is composed of six light $\varnothing 8 \si{\milli\meter}\!\times\!2 \si{\milli\meter}$ carbon-fiber tubes forming two equilateral triangles that are disposed in opposition to form a  hexagram star. 
Motors are placed slightly outward of each vertex, resulting in a diagonal distance between propeller axes of $370 \si{\milli\meter}$.
This unusual hexagram design provides high rigidity and low weight (see \tabref{tab:ro_platform_weight_list}). 
The inner part of the star forms a generously sized hexagon, not spanned by the propellers and hence not interrupting the main airflow, where the main body with all other hardware resides.
3D-printed parts and regular M3 screws hold together the tubes and the rest of the components.
The control, power, and communication hardware is above the star plane, with the battery at the top. 
This provides two advantages for an agile motion: 
First, the \gls{com} is located high, allowing for rapid and easy destabilization from the hovering stance;
Second, the limb, which is attached below the base, can have the first joint's axis close to the body's propeller plane and \gls{com}, thereby improving the dynamic behavior of the assembly in the face of rapid tilting of the base.
Four small, mechanically fusible, and easily replaceable 3D-printed pieces are fixed to the carbon tubes to act as ports to fasten the lower body (\figRef{fig:ro_platform_overview}-right and section \ref{subsec:limb}).
The platform can be manufactured using only a 3D printer and common tools such as a soldering iron, metric Allen keys, and wrenches.

\begin{table}[t]
  \caption{Weights and approximate prices of the platform's components.}
  \label{tab:ro_platform_weight_list}
  \begin{center}
    \begin{tabular}{@{}l c c@{}}
      \toprule
      & \textsc{Weight $[\si{\gram}]$} & \textsc{Cost (approx.) [\texteuro]} \\
      \midrule
      \textsc{Platform} \\
      \midrule
      \emph{Structural parts} \\
      3D printed parts & 544 & 20.00 \\ 
      Carbon fiber tubes & 82 & 41.00 \\
      \midrule
      \emph{Drive system} \\
      Motors & 348 & 165.00 \\
      ESCs   &   54 & 160.00 \\
      Propellers  &   36 & 23.00 \\
      \midrule
      \emph{Others} \\
      Battery 6S 3.0\,Ah 60C   &  498 & 83.00 \\ 
      Power module PM03      & 80 & 55.00 \\
      Pixhawk V5X           & 87 & 450.00 \\
      NUC7-i7-DNKE & 232 & 1050.00  \\
      Misc. &   151 & - \\
      \midrule
      \textbf{\textsc{Total platform}} &  2112 & 2047.00 \\     
      \bottomrule
    \end{tabular}
  \end{center}
\end{table}

\begin{table}[t]
  \caption{Weights and approximate prices of the 2 \gls{dof} limb's  components.}
  \label{tab:ro_arm_weight_list}
  \begin{center}
    \begin{tabular}{@{}l c c@{}}
      \toprule
      & \textsc{Weight $[\si{\gram}]$} & \textsc{Cost (approx.) [\texteuro]} \\     
      \midrule
      \textsc{2 \gls{dof} limb} \\
      \midrule
      \emph{Legs \& ODRI boards} \\
      3D printed parts & 130 & 3.0 \\ 
      Carbon fiber tubes & 37 & 14.00 \\
      ODRI micro driver & 30 & 50.00 \\
      ODRI master board & 18 & 40.00 \\
      \midrule
      \emph{Links} \\
      1st Link & 261 & 300.00 \\
      2nd Link & 139 & 300.00 \\
      End effector (tail) & 127 & 1.00 \\
      End effector (leg) & 25 & 1.00 \\
      \midrule
      \textbf{\textsc{Total limb (tail) }} & 742 & 709.00 \\     
      \bottomrule
    \end{tabular}
  \end{center}
\end{table}

The Borinot flying platform holds different subsystems:
\begin{enumerate*}
  \item the power and drive systems, comprised by the motors, the \glspl{esc} and the battery;
  \item the flight controller, which manages functionalities related to flying such as the state estimation, the communication with a radio controller, handling safety flight modes, and interfacing with the \glspl{esc}; and
  \item the onboard computer that is used to run the algorithms that operate the robot.
\end{enumerate*}

\subsubsection{Power \& Drive system}

\begin{figure}[t]
  \centering
  \includegraphics[width=\columnwidth]{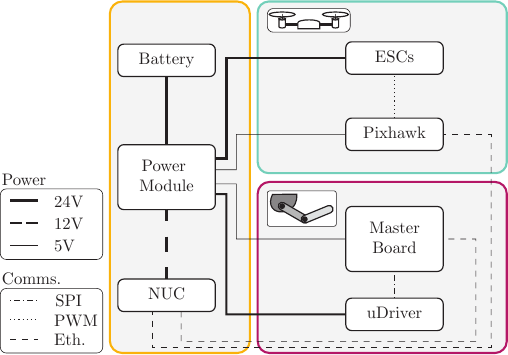}
  \caption{Borinot's power distribution scheme and communication between the different components. The power module is powered by the $6S$ $24\si{\volt}$ battery. From it, the power is properly converted and distributed to each component. 
  The \glspl{esc} and the limb's uDrivers are powered directly at $24\si{\volt}$. The NUC onboard computer is powered at $12\si{\volt}$ while the Pixhawk and the arm master board are powered at $5\si{\volt}$. 
  The onboard PC communicates with the Pixhawk through an ethernet connection for fast data exchange during the operation, and through USB to access the PX4 command-line tool. 
  Additionally, The NUC and the master board communicate via a second ethernet adapter.}.
  
  \label{fig:ro_power_distribution}
\end{figure}

To achieve a high \gls{twr} we have to minimize weight (see \tabref{tab:ro_platform_weight_list}) and maximize thrust.
Regarding thrust, the range of available \gls{bldc} motors for multi-copters has increased significantly in recent years, leading to motors with special characteristics tailored to different applications (\gls{fpv}, aerial photography and cinema, surveillance, etc.). 
Our platform's dimensions limit the propellers' size to $7$ inches.
Thus, we target motors mainly used within the \gls{fpv} community, typically using small propellers.
The selected \textsc{TMotors F90-1300KV} \cite{tmotor_datasheet}, powered by a $6\text{S}$ Li-Po battery (nominal $22.2\si{\volt}$) and equipped with a 7-inch propeller with 4.2 inches pitch, can produce up to $16.1\si{\newton}$ of thrust. 
This results in a total thrust of $96.6\si{\newton}$ for the hexacopter.
More details on the performance of this motor-propeller set are provided in \secref{subsec:exp_motor_tests}. 

The motor selection highly restricts the choice of the \glspl{esc} and the battery.
The selected \gls{esc} is the \textsc{TMotor F35A 6s}, a drone racing component that allows operating the motor with a $6S$ battery and a fast throttle response.
As for the battery, we have selected a $6\text{S}$ unit with $3\si{\ampere\!\,\hour}$ and a discharge rate of $60\text{C}$.
This provides room to continuously operate the robot at $90\%$ of its throttle capacity, and having burst loads up to $100\%$. 
For reference, Borinot with a 2DoF limb has a \gls{twr} of 3,5: this means that its motors are at $100/\textrm{\gls{twr}}=28,6\%$ of throttle capacity during hovering.

\subsubsection{Flight controller}

The flight controller selection is usually tied to the type of firmware that we want to use.
We have opted for a unit compliant with the PX4 flight stack (see the justification of this choice in \secref{subsec:so_px4}).
We have selected the \textsc{Holybro Pixhawk 5X} because of its fast Ethernet communication with the main onboard computer.
The unit includes three IMUs, two barometers, and two GPS ports, has built-in communication with the radio controller and other inputs such as motion capture, and can drive up to 16 servos and \glspl{esc}.

\subsubsection{Onboard computer}
Borinot is equipped with an Intel NUC7-i7-DNKE, which mounts an Intel i7-8650U 4-cores CPU, a clock frequency of $1.9\si{\giga\hertz}$ and $32\si{\giga\byte}$ of DDR4 RAM.
This computer comfortably runs computation-intensive controllers, such as \gls{mpc}.
The CPU benchmarks 6300 at cpubenchmark\footnote{\url{https://www.cpubenchmark.net/cpu_lookup.php?cpu=Intel+Core+i7-8650U+\%40+1.90GHz\&id=3070}}, a website used in \cite{foehn_AgiliciousOpensourceOpenhardware_2022} to quantify the computing power of research multicopters. 
In this sense, Borinot compares favorably to Agilicious \cite{foehn_AgiliciousOpensourceOpenhardware_2022}, with a similar \gls{twr} but a CPU mark of 6300 vs. 1343 for Agilicious.

\subsection{Lower body: limbs and landing gear}
\label{subsec:limb}

The lower body (\figref{fig:ro_platform_overview}, right) comprises the landing gear and the limb or limbs. 
The assembly is attached to the carbon tubes of the upper body through 4 attachment ports so that different lower-body designs can be accommodated. 
The landing gear is a light structure made of carbon tubes whose mission is to protect the limbs upon landing. 
This section is, therefore, devoted to limb design.

To comply with the requirements in \secref{subsec:int_requirements} that affect the limb, we take advantage of the \gls{odri} ecosystem \cite{grimminger_OpenTorqueControlledModular_2020}.
This initiative proposes an open-source robotic actuator (electronic and mechanical parts) to build affordable agile legged robots such as quadrupeds or bipeds.
Adopting this technology allows us to work with different kinds of limbs attached to the same upper body.
A particular limb design can be selected according to the kind of task or research to conduct.
We judge this to be a better solution than having one universal  limb with many \glspl{dof}, which due to its weight, would undermine the final \gls{twr} and thereby the agility potential of Borinot.

The \gls{odri} actuator consists of a drone motor able to produce 0.3Nm of torque, a rotary encoder at the motor axle, and two pinion-belt 1:3 reduction stages. This results in a 1:9 gear ratio with an output torque of 2.7Nm. 

The electronics for the limb are composed of one \gls{odri} master board, which handles the communication with the onboard computer, connected to up to six uDrivers, which can drive two motors each. The uDrivers can perform torque control and variable-impedance control.

\begin{figure}[t]
  \centering
  \includegraphics[width=0.5\columnwidth]{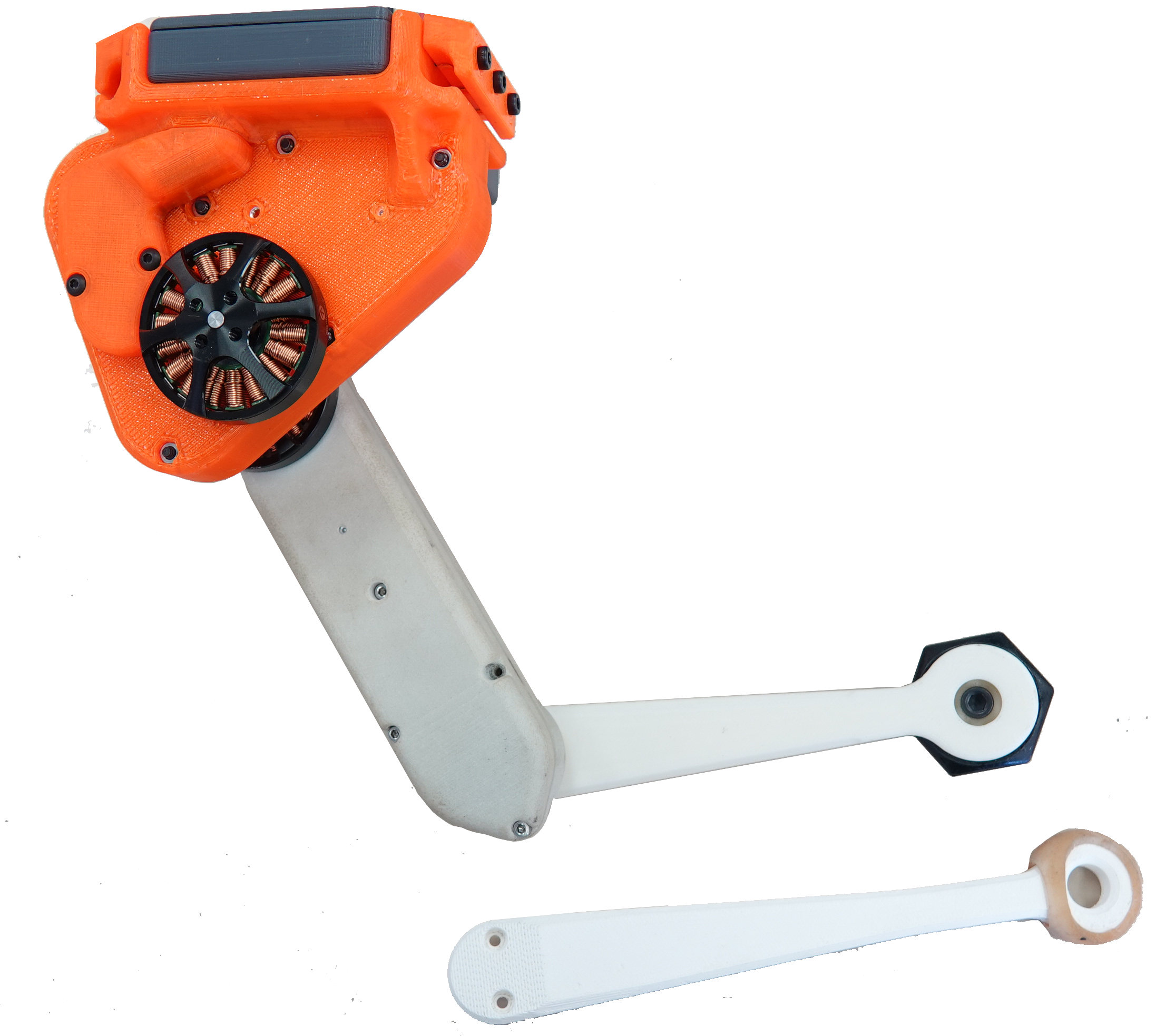}
  \caption{Borinot's $2$-\gls{dof} limb, with a 100g weight (an M24 nut) on its end-effector to be used as a tail in agile flying locomotion. An alternative terminal link with a rubber boot is shown below, to be used as a leg.}
  \label{fig:ro_arm_design}
\end{figure}

In this work, we implement a planar $2$-\gls{dof} limb (see \figref{fig:ro_arm_design}), which allows us to investigate many of the main concepts of agility while keeping the weight low.
We have designed a new limb's base link that contains the same hardware parts as the \gls{odri} actuator at a smaller footprint.
We used a Solo12 leg link directly from \gls{odri} for the second link.
The third or terminal link can be selected depending on the application (\figref{fig:ro_arm_design}). 
When using the limb as a leg, we mount a regular Solo-12 terminal link with a rubber boot to maximize contact friction. 
When using it as a tail or arm, we mount a custom 3D-printed link on which a heavy object can be attached to increase the inertia of the end effector.

\section{Borinot's software}
\label{sec:software}

A robot intended for research purposes should come with the necessary software that enables some basic functionalities such as low-level controllers, basic state estimation, communication protocols, and other essential tools. 
However, developing these functionalities from scratch can be a resource-intensive task for laboratories.
We opted not to take this approach, and we sought hardware that was compatible with software that already provided these functionalities, requiring only some modifications.

\begin{figure}[t]
  \centering
  \includegraphics{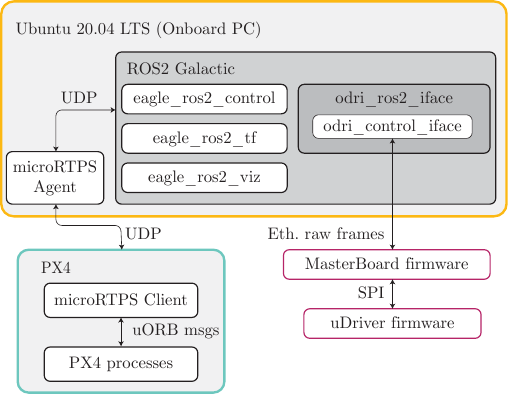}
  \caption{Borinot's basic software layout. The NUC runs on Ubuntu 20.04 LTS and has the Galactic version of ROS2 as the middleware to integrate all the robot's modules. The Pixhawk runs a modified version of the PX4 firmware. It leverages the implementation of the microRTPS agent and client to connect with the NUC and communicate with the different ROS2 nodes. The exchange of the limb's data is done directly from a ROS2 node that uses an ODRI C++ library to communicate with the master board.}
  \label{fig:software_layout}
\end{figure}

In \figref{fig:software_layout} we show the different software packages that are used in Borinot and how they are distributed among its hardware (NUC as the main processor, Pixhawk as the flight controller, and \gls{odri}'s masterboard as the limbs' controller).
They are described in the following sections.

\subsection{Robot Operating System}
We have chosen ROS 2\footnote{\url{www.ros.org}} as the middleware to run in the onboard NUC computer. 
It enables us to easily integrate the different modules contained in Borinot.
The use of ROS is three-fold. 
First, it is used as a tool to easily exchange data between different modules. 
For example, to send \gls{mocap} poses to PX4 to be fused with the IMU, or to gather the estimated state from PX4.
Second, it is used as the interface for the algorithms that we are researching, which for the sake of modularity are implemented in separate stand-alone libraries.
And third, ROS is used to implement tools that are necessary to operate the robot, such as visualization tools or \glspl{gui} to send commands to the robot.
We have gathered all the ROS software tools related to Borinot as ROS 2 packages and placed them in a single repository at \texttt{\url{https://github.com/hidro-iri/eagle_ros2}}: 
\begin{enumerate}
    \item \textbf{\texttt{eagle\_ros2\_bag}}: tools to analyze ROS2 bags.
    \item \textbf{\texttt{eagle\_ros2\_bringup}}: launch and configuration files.
    \item \textbf{\texttt{eagle\_ros2\_control}}: a C++ library with base classes that can be derived to implement custom controllers. It already contains two controllers: a controller based on the off-the-shelf PX4 controllers, and the MPC controller that we used to create the experiments in this paper (described in \secref{sec:control_architecture}). 
    \item \textbf{\texttt{eagle\_ros2\_identification}}: tools to identify the motor-propeller system. That is, the algorithms to run the experiments from \secref{subsec:exp_motor_tests} and a ROS2 node to gather data from the thrust stand.
    \item \textbf{\texttt{eagle\_ros2\_interfaces}}: These are messages and services for interfacing with other packages. 
    \item \textbf{\texttt{eagle\_ros2\_tf}}: a ROS2 base class node that collects the PX4 and ODRI data composing the state of the robot, \ie, poses, joint configuration and their corresponding velocities. 
    Since controllers in the package need the robot state, they must all derive from this class. 
    Additionally, this package contains a node that collects \gls{mocap} poses and does the necessary frame conversions to send them to PX4 for their fusion within its \gls{ekf}.
    \item \textbf{\texttt{eagle\_ros2\_viz}}: this package contains two \glspl{gui}: a state visualization \gls{gui} and another one to send commands to the robot.
\end{enumerate}

\subsection{Flight controller firmware: PX4}
\label{subsec:so_px4}
The PX4 firmware provides a comprehensive suite of functionalities for conducting research with multicopters.
In addition to the necessary auxiliary machinery required to operate a multicopter-based robot, such as radio systems, flight modes (including safety flight modes), and configuration tools, PX4 includes a state estimator that is crucial for accurate and reliable system control.
The state estimator%
\footnote{https://github.com/PX4/PX4-ECL}
facilitates the fusion of a wide range of sensors, including \gls{mocap}, and provides researchers with a reliable foundation to focus on other areas of the robot, such as control.
PX4 also provides a rigorously tested set of flight controllers, which can be used as recovery/safety modes or backup when developing and testing other control algorithms.
Furthermore, the microRTPS client in PX4 along with the Ethernet communication in the flight controller allows for high rate data exchange with ROS2.
Besides, as PX4 is an open-source project that has been adopted by a wide community including the industry, commercial application and research makes it suitable for an open-source multicopter-based robotic platform, such as Borinot.

However, the off-the-shelf PX4 firmware lacks some features that are necessary for our control algorithms, requiring modifications.
Notably, direct access to the \gls{esc} inputs, which is crucial for most control algorithms, is not available.
Additionally, the microRTPS client provided by PX4 does not allow for fast data sending to the onboard computer.
To address these limitations, we developed modifications based on PX4 firmware version 1.12. 
These modifications include the creation of a new flight control mode that operates independently of any existing control modes and the introduction of a new uORB message in PX4 that intervenes only when the new flight mode is active.
By completely isolating the direct control pipeline from PX4's existing pipeline of controllers, our modifications ensure that the latter remains untouched and ready to be activated in the event of an emergency situation with experimental algorithms. 
We also modified the microRTPS client to enable higher-rate sensor data sending.

\subsection{ODRI software}
The ODRI initiative provides firmware that runs on both the micro driver and the master board. 
It also provides a C++ library with an API to receive data from the master board.
However, it does not provide a ROS2 package that enables sending and receiving data to/from the master board.
To do that, we have implemented a ROS2 node that uses the C++ API to communicate with the master board. 
This node contains the necessary publishers and subscribers to operate an ODRI-based robot from ROS2.

\section{Control architecture}
\label{sec:control_architecture}

\subsection{Overview}
The control requirements for an agile loco-manipulator have been discussed in \secref{sec:conditions-agile-robots}.
These conditions can be summarized as whole-body, dynamics, torque-level, optimality, and predictiveness, using contacts and flight.
Among other possible options, a good candidate for conceiving such controllers is the \gls{mpc}, a model-based paradigm that uses \gls{oc} at its core.
Solving a torque-based, whole-body dynamics \gls{ocp} we are able to satisfy the requirements for dynamics, optimality, and maneuvering. 
Incorporating contacts yields a capable architecture for controlling any agile loco-manipulator.
The dynamics-level \gls{mpc} paradigm is well known and widely used by the legged robotics community \cite[and many others]{Carpentier_2018}, and also in agile flying \cite{foehn_AgiliciousOpensourceOpenhardware_2022}, but has been seldom explored for \glspl{uam} \cite{Ollero_2022}.
We, therefore, find it relevant to provide a basic introduction in the form of a preliminary yet functional design.

\subsection{Features and limitations}
What we present here is a fairly simple \gls{mpc} approach to controlling Borinot, which serves two purposes.
Firstly, as an illustration of a valid methodology for controlling such robots. 
And secondly, as a means to experimentally demonstrate some of the abilities of Borinot as an agile flying platform.
However, given the complexity and variety of the hybrid motion modes Borinot is designed to handle, the presented method lacks some features that will need to be developed in future improvements.
First, it does not account for contacts. 
This means that we cannot use it in the experiments section to demonstrate locomotion with contacts.
And second, it does not consider the task specification inside the control loop (see \figref{fig:control_architecture}).
This means that it suffers from some lack of accuracy, especially in positioning the arm's end effector, which is important for manipulation. 
These two issues are not inherent to the \gls{mpc} paradigm but to our preliminary implementation, and should not represent a fundamental impediment to designing better \gls{mpc} controllers.

\subsection{Architecture}
\begin{figure}[t]
  \centering
  \includegraphics{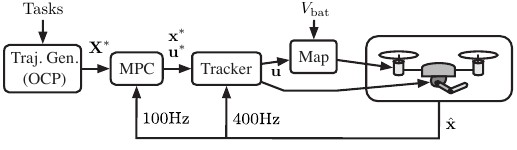}
  \caption{The Borinot control architecture used in our experiments is based on expressing tasks as residuals to be minimized in whole-body dynamics \gls{ocp}. 
  By solving the \gls{ocp}, we obtain an optimal state trajectory that is subsequently used as a reference for a \gls{mpc} controller, which runs at $100\si{\hertz}$. 
  The optimal \gls{mpc} solution for state and control, denoted as $\bfx^*, \bfu^*$, is used as the reference state and feed-forward values in a lower-level controller running at $400\si{\hertz}$.
  The produced thrust commands are mapped onto motor commands and sent to the \gls{esc}s via PX4, while limb torques are fed directly to the \gls{odri} master board for variable-impedance control.}
  \label{fig:control_architecture}
\end{figure}

The control architecture is depicted in \figref{fig:control_architecture}.
It consists in the first place of a whole-body dynamics \gls{ocp} that is solved offline.
The obtained optimal state and control trajectories already capture most of the conditions for agility: optimality, dynamics, maneuvering, and redundancy, and can provide accurate open-loop solutions for contact-less manipulation tasks.
The state trajectory $\bfX^*$ is used as a reference by a whole-body dynamics \gls{mpc} controller, running at $100\si{\hertz}$.
This controller is fed with state feedback $\hat\bfx$ from the state estimator in PX4 and produces new state and control trajectories.
Next, we have a tracking controller to compensate for model mismatches, which runs at  $400\si{\hertz}$.
Finally, the resulting limb torques, angles and velocities are sent to the \gls{odri} variable-impedance controller, and propeller thrusts are mapped to motor commands, which are sent to the \gls{esc}s.
In the following, we give more details about each module in the architecture.

\begin{figure}[t]
  \centering
  \includegraphics[width=\columnwidth]{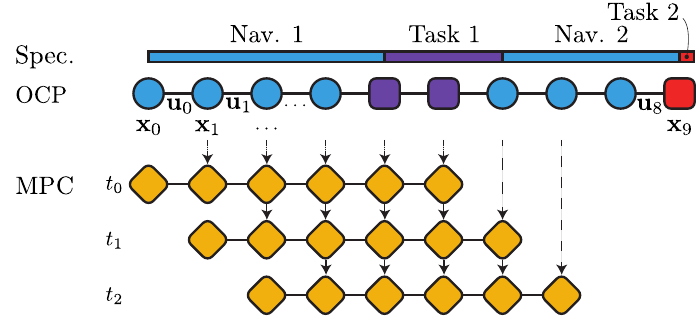}
  \caption{Process to go from a mission specification to the assembly of the \gls{mpc} controller. At the top, we have the mission specification, composed of navigation and task phases.
  These phases are encoded in terms of residuals in the offline \gls{ocp} (blue, purple and red nodes). Then, the optimal state trajectory, solution of the \gls{ocp}, is used as a reference to build the successive \glspl{ocp} of the \gls{mpc} controller (yellow nodes).}
  \label{fig:control_mission_spec}
\end{figure}

\subsubsection{From mission to \gls{mpc}}
In  \figref{fig:control_mission_spec} we show the process of going from the specification of a mission to assembling the successive \glspl{ocp} to be solved by the \gls{mpc} controller.
A robot mission is composed of phases that can be of \emph{task} type or of \emph{navigation} type.
Tasks, which are colored in purple and red, can be for example to bring the robot to a specific location, place the end-effector at a precise position, or pass through a narrow window. 
Navigation phases are placed before tasks and are used to prepare a feasible and optimal maneuver so that the task can be accomplished.

The mission's timeline is discretized into a predefined set of nodes linked by control actions.  
A residual is associated with each node: task nodes have strong residuals, and navigation nodes have weak regularization residuals.
The set of all residuals forms the following \gls{ocp}, which is solved offline once:
\begin{align}
  \begin{aligned}
     \bfX^*,\bfU^* =~ 
     & \underset{\bfX,\bfU}{\min} &  & \sum_{k=0}^{N-1}l^{\textrm{OCP}}_k(\bfx_k, \bfu_k) + L(\bfx_N)                                          \\
     & \text{s.t.}                &  & \bfx_{k+1} = f(\bfx_k, \bfu_k) \,,                                    ~~k\in [0, N-1] \,, \\
     &                            &  & \bfx_{0} =  \overline{\bfx}_0                                    \,,                     \\
     &                            &  & \underline{\bfu} \leq \bfu_k \leq \overline{\bfu}  \,,                ~~~~~~~~~k\in [0, N-1] \,, \\
  \end{aligned}
  \label{eq:control_architecture_docp}
\end{align}
Here, $\bfX=\{\bfx_k\}$ and $\bfU=\{\bfu_k\}$ are the state and control trajectories specified for $N$ nodes.
States comprise the robot's configuration and its velocity, $\bfx_k\in SE(3)\times\bbR^{6+2n_\textrm{joints}}$, while controls consist of propeller thrusts and limb joint torques, $\bfu_k\in\bbR^{n_\textrm{props}+n_\textrm{joints}}$.
The function $l^{\mathrm{OCP}}_k(\cdot)$ is the residual cost for the node $k$, and $L(\cdot)$ is the terminal cost.
The function $f(\cdot)$ is the integration of the robot's forward dynamics over the discretization step assuming constant control forces and torques $\bfu(t)=\bfu_k$, $\overline{\bfx}_0$ is the initial state and $\{\underline\bfu,\overline\bfu\}$ are the actuation bounds.
Non-actuation limits such as joint limits are incorporated as quadratic barriers in the cost function $l^{\mathrm{OCP}}_k(\cdot)$.
We solve \eqref{eq:control_architecture_docp} using control-limited  \gls{fddp} techniques as in \cite{mastalli-20-crocoddyl,martisaumell-2020-squashbox}.

The optimal state trajectory $\bfX^*$ of \eqref{eq:control_architecture_docp} is used as a reference $\bfX^{\textrm{ref}}$ for the \gls{mpc} controller.
This controller successively solves \glspl{ocp} like \eqref{eq:control_architecture_docp}, of shorter horizon and initial condition $\bfx_0=\hat\bfx$ taken from the state estimator.
These \gls{ocp}s consider, at each node $k$, a residual that follows the reference trajectory while minimizing actuation effort:
\begin{align}
l^{\textrm{MPC}}_k(\bfx_k, \bfu_k) = \cfrac{1}{2} ||\bfx_k \ominus \bfx^{\textrm{ref}}_k ||_{\bfW_\bfx}^2 + \cfrac{1}{2} || \bfu_k ||_{\bfW_\bfu}^2 \,, \label{equ:mpc}
\end{align}
where $\ominus$ produces an error vector in the tangent space of the manifold of $\bfx_k$ \cite{sola-18-Lie}.
Here, $\bfW_\bfx$ and $\bfW_\bfu$ are constant pre-defined weights and, in consequence, the distinction between navigation and task phases is no longer present in the \gls{mpc} problems \eqref{equ:mpc}.
These problems are solved with the same \gls{fddp} techniques \cite{martisaumell-2020-squashbox},  allowing on-board real-time execution.
For further details we refer the reader to the \emph{Rail-MPC} controller in  \cite{martisaumell2021fullbody}, whose name comes from the fact that the offline trajectory $\bfX^{\textrm{ref}}$ serves as a rail in which the \gls{mpc} controller runs.

\subsubsection{Tracking controller}
The result of the \gls{mpc} is fed to the tracking controller.
This controller receives two inputs from the \gls{mpc}: feed-forward control commands $\bfu^*$ (propeller thrusts and limb torques), and  whole-body state references $\bfx^*$, which are used in a proportional-derivative feedback loop.
This is basically an adaptation of the controllers proposed in, \eg, \cite{lee_GeometricTrackingControl_2010} or \cite{tognon_dynamic_2017}, where we use a full-state reference trajectory from the \gls{mpc} instead of a flat outputs reference trajectory.

\subsubsection{Final motor commands}
The output $\bfu$ of the tracking controller is a vector of thrusts and limb torques. 
Limb torques, desired angles, velocities, and impedance gains are sent directly to the \gls{odri} variable-impedance controllers. 

Regarding thrusts, our \glspl{esc} implement open-loop trapezoidal control for sensorless \gls{bldc} motors, which provides no direct control over the motor's thrust or speed%
\footnote{There exist \glspl{esc} that implement closed-loop speed control. However, these are less common, more expensive, and do not always align with the desired specifications. See, \eg, \cite{franchi_AdaptiveClosedloopSpeed_2017}.}
\cite{yedamale_AN885ApplicationNote_}.
To overcome this, the thrusts are fed to a mapping module that converts thrusts into motor commands. 
We implement this mapping by fitting a 3rd-degree polynomial surface that takes into account battery voltage, input command, and produced thrust.
Data for the fitting came from our motor-propeller calibration workbench \cite{thrust_stand}. 
The performance of this mapping is evaluated in \secref{subsec:exp_motor_tests}.

\section{Experiments}
\label{sec:experiments}

There are different things that we aim at validating regarding Borinot's capabilities.
First, in \secref{subsec:exp_motor_tests} we want to gain insight into the performance of its hexacopter platform.
We are interested in quantitative details during a normal near-hovering operation and during a demanding scenario.
Quantities such as its thrust capacity, its power delivery, or its current consumption allow us to validate the component selection and to know where its limits are in terms of performance.

Then, we test the robot for the different motion modes involved in loco-manipulation, as explained in \secref{subsec:types_agile_locoman}, using the limb as a tail in \secref{subsec:exp_flying_locomotion}, as an arm in \secref{subsec:flying_manipulation}, and as a leg in \secref{subsec:exp_aerial_contact_locomotion}.
For the first two experiments, we use the control architecture presented in \secref{sec:control_architecture} running onboard the NUC i7 computer.
We use $35$ nodes and a node separation of $20\si{\milli\second}$, giving an \gls{mpc} horizon of $0.7$ seconds.
The last experiment involves contacts and jumps, and for the sake of simplicity, we resort to open-loop commands and the assistance of a guide rail.

\subsection{Motor-propeller tests} 
\label{subsec:exp_motor_tests}

\begin{figure*}[t]
  \centering
  \includegraphics[width=\textwidth]{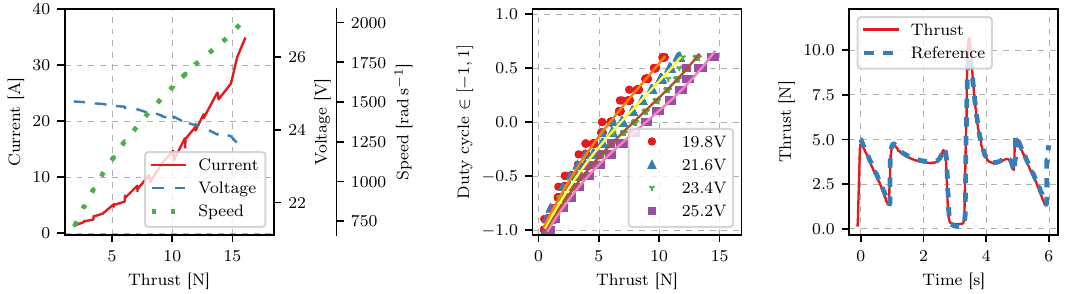}
  \caption{
  Motor experiments in the thrust stand.
  \emph{Left:} 
  The platform motor's thrust vs. speed curve (dotted) with the $7 \times 4.2$ propeller. 
  We tested the motor with a fully charged $6S$ battery. 
  The motor can deliver up to $16.1\si{\newton}$ of thrust, which leaves us with a total theoretical thrust for the platform of $96.6\si{\newton}$ and a \gls{twr} value of $3.5$. 
  Notice that if we consider the platform alone, the \gls{twr} increases to $4.7$.
  Motor current consumption and battery voltage (solid and dashed, respectively). We show an experiment with a fully charged battery, \ie, $25.2\si{\volt}$. 
  The motor reaches a current of $35\si{\ampere}$ at full throttle, delivering $16.1\si{\newton}$. 
  To keep the integrity of all the components of the robot, we should be below $30\si{\ampere}$ per motor and only go over this point for short periods.
  \emph{Center:}
  Thrust vs. \gls{esc} command map for different battery voltage levels. For this specific experiment, we have kept the voltage leveled using a DC power supply. 
  Lines show the duty cycle values resulting from the evaluation of the fitted 3rd-degree polynomial with each thrust and voltage value.
  \emph{Right:}
  Mapping accuracy, comparing the desired thrust (dashed line) with the real thrust using the mapping, which is measured by the thrust stand.
  }
  \label{fig:exp_motors}
\end{figure*}

To get an estimate of the \gls{twr} and its implications in terms of energy consumption and autonomy, we have tested a single motor in a test bench for motor-propeller systems \cite{thrust_stand}.

In \figref{fig:exp_motors}-left, we show the thrust-speed curve of the platform motor with the $7 \times 4.2$ propeller.
The motor can deliver $16.1\si{\newton}$ of thrust, which translates into $96.6\si{\newton}$ of theoretical thrust for the platform.
This means a \gls{twr} of $4.7$ when using only the platform and $3.5$ with the platform and the limb.

In terms of energy consumption, \figref{fig:exp_motors}-left also shows the current that the motor draws to achieve different thrusts.
The hovering current consumption is at $30\si{\ampere}$ for the case with the limb and $20\si{\ampere}$ for the platform alone ($5\si{\ampere}$ and $3\si{\ampere}$ per motor to achieve a total thrust of $28\si{\newton}$ and $20\si{\newton}$, respectively).
Considering the $3\si{\ampere\hour}$ battery, this consumption leads to a theoretical hovering autonomy of $4\si{\minute}$ and $8\si{\minute}$, respectively.
This autonomy is enough for doing several attempts of an agile trajectory, which normally take less than a minute each.

The mapping from the desired thrust to motor command depending on battery voltage is shown in \figref{fig:exp_motors}-center.
Its accuracy and dynamic response are shown in \figref{fig:exp_motors}-right, where we observe a tight match between desired and achieved thrusts, with a sharp response to transients.

\subsection{Flying locomotion: limb as a tail}
\label{subsec:exp_flying_locomotion}

\begin{figure*}[ht]
  \centering
    \includegraphics[width=0.135\textwidth]{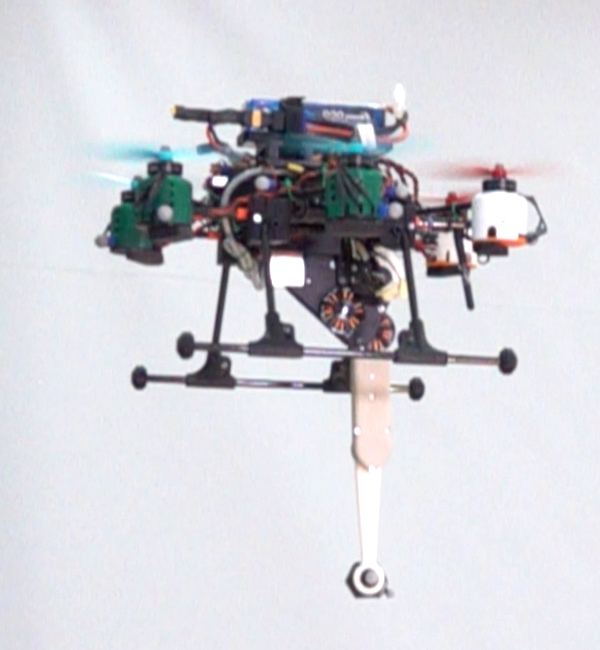}
    \includegraphics[width=0.135\textwidth]{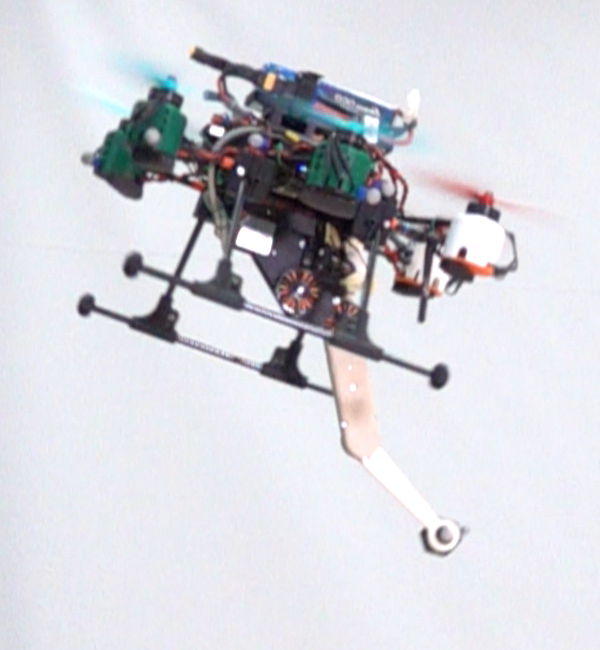}
    \includegraphics[width=0.135\textwidth]{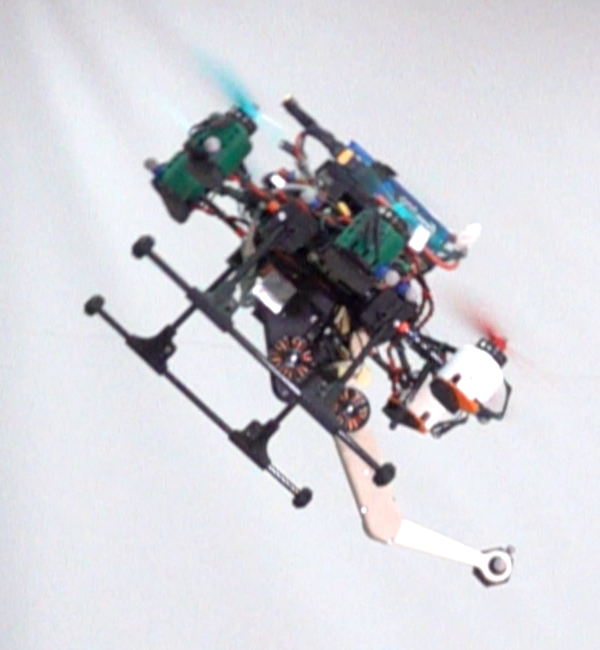}
    \includegraphics[width=0.135\textwidth]{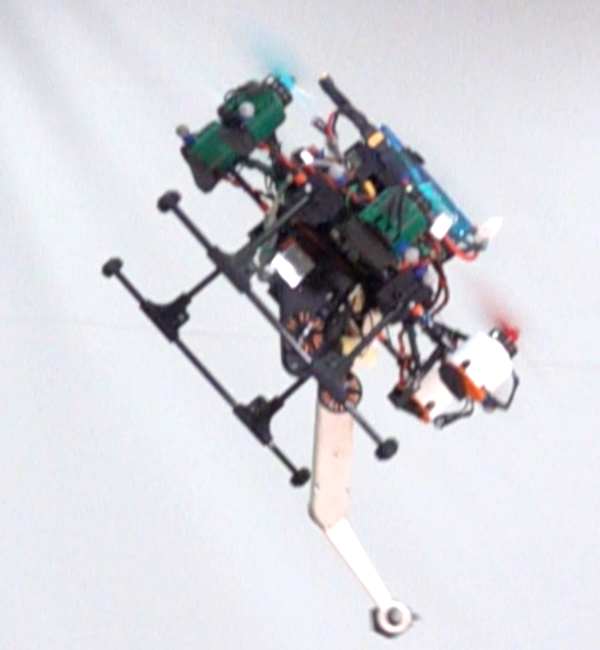}
    \includegraphics[width=0.135\textwidth]{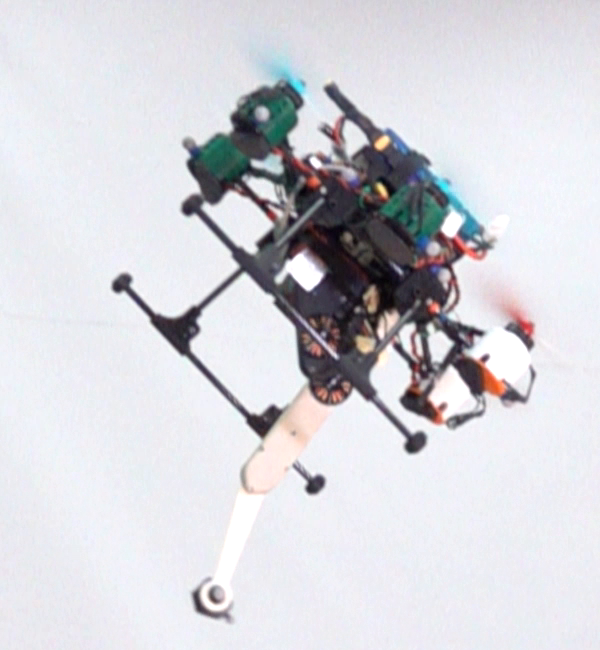}
    \includegraphics[width=0.135\textwidth]{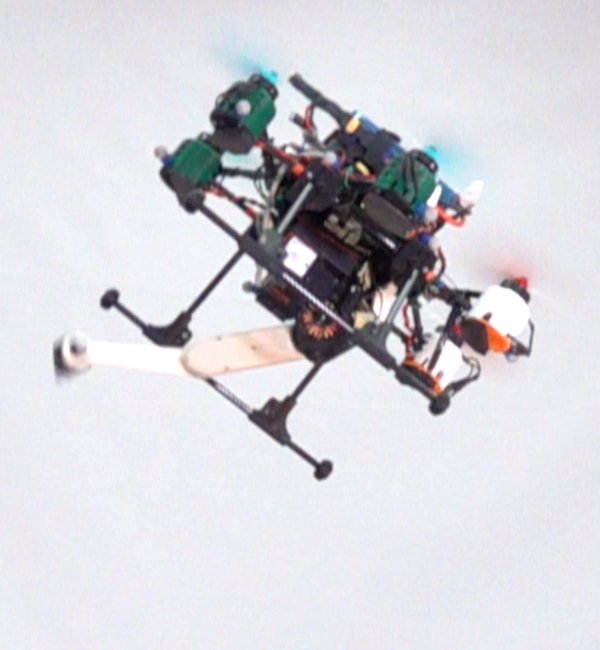}  
    \includegraphics[width=0.135\textwidth]{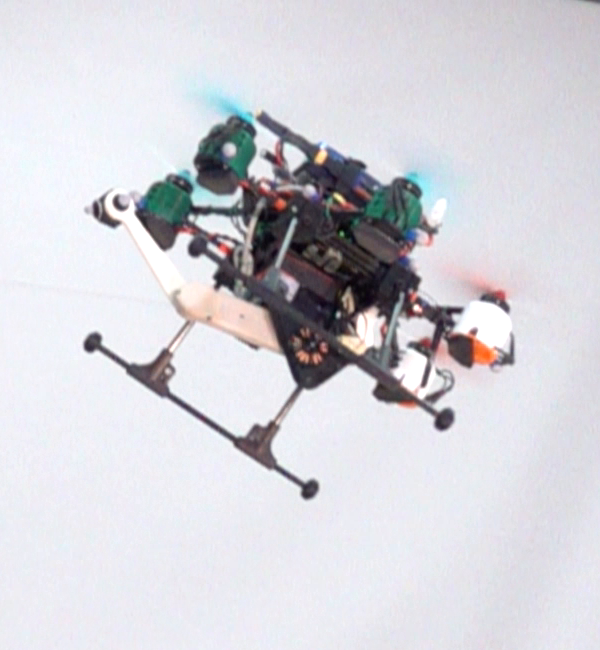}
  
  \vspace{1ex}
  
    \includegraphics[width=0.135\textwidth]{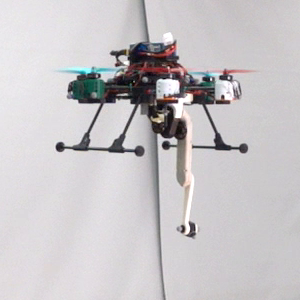}
    \includegraphics[width=0.135\textwidth]{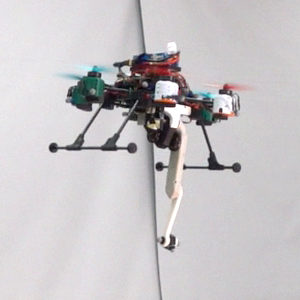}
    \includegraphics[width=0.135\textwidth]{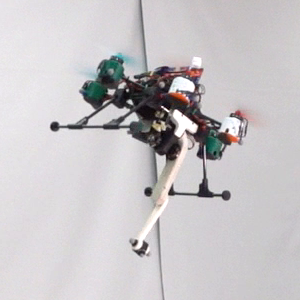}
    \includegraphics[width=0.135\textwidth]{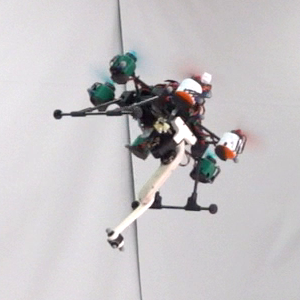}
    \includegraphics[width=0.135\textwidth]{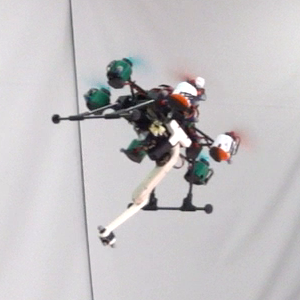}
    \includegraphics[width=0.135\textwidth]{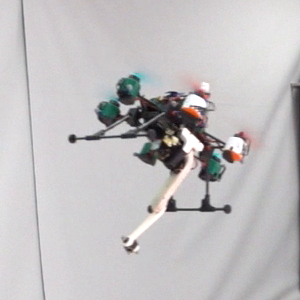}  
    \includegraphics[width=0.135\textwidth]{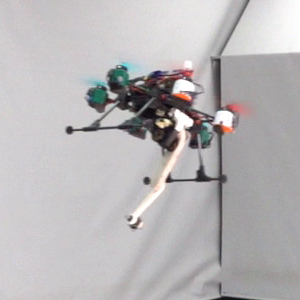}
  
  \caption{Image sequences of the different agile flying trajectories showing the initial acceleration maneuver (around 0.5\,s). 
  \emph{Top:} fast displacement in the sagittal plane, showing dynamic tail assistance in the form of rapid limb movements. 
  \emph{Bottom:} fast displacement in the coronal plane, showing static tail assistance consisting on centering the overall \gls{com}.
  To get a better understanding of the trajectories, we encourage the reader to see this paper's video.
  }
  \label{fig:sequence-displacements}
\end{figure*}

The first set of experiments illustrates the use of the limb as a tail to participate in flight dynamics.
The setup is as follows.
We define the body axes of Borinot as X-forward, Y-left, and Z-up, and identify the XZ and YZ planes respectively as the sagittal and coronal planes.
We mount a 2\gls{dof} planar tail so that its motion occurs only in the sagittal plane. 
We thus expect it to contribute differently to trajectories evolving in the sagittal and coronal planes.
A weight of 100g is attached to the endpoint of the tail to increase its inertia and therefore its impact on the overall dynamics (see \figref{fig:ro_arm_design}).

We specify two trajectories, namely \emph{sagittal displacement} and \emph{coronal displacement}, which are rapid movements along the X and Y directions, respectively (see \figref{fig:sequence-displacements} for the initial acceleration phase of these trajectories).
Both share the same fundamental task of moving from an initial hovering position to a waypoint placed at $5\si{\meter}$ distance using a time interval of 2\,s, remaining there for $600\si{\milli\second}$, then returning to the starting point also in 2\,s, and finally hover.
The tail configuration at all waypoints (initial, intermediate, and final) is set to be fully stretched down.

The tail's influence on the platform dynamics can be of different nature.
On the one hand, the tail can be moved to adjust the overall \gls{cog} and reduce the moment of inertia, enabling the platform to tilt more easily. 
We refer to this as \emph{static tail assistance}, since this adjustment is usually done at low velocities.
On the other hand, a high torque can be applied at the tail's joints, producing a reaction torque at the platform and resulting in a rapid change in inclination.
We refer to this as \emph{dynamic tail assistance}, as the torque considerably alters the angular momentum.
In the present setup, dynamic assistance is not practicable in the coronal movements since the tail can only apply torques in the sagittal plane.

\begin{figure}[t]
  \centering\includegraphics{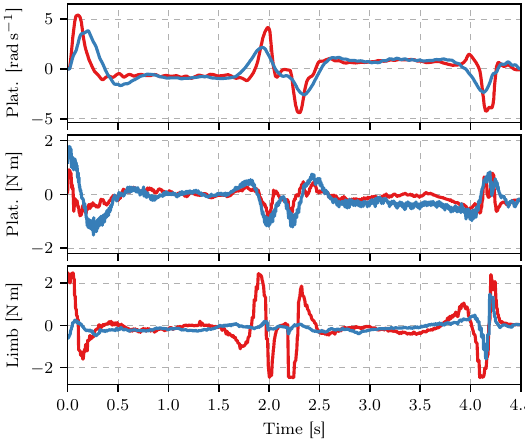}
  \caption{Limb as a tail and its influence on the overall motion for the sagittal (red) and coronal (blue) trajectories. 
  \emph{Top:} platform's angular speed in the respective planes. 
  \emph{Center:} torque applied to the platform only due to propeller thrusts. 
  \emph{Bottom:} torque of the first joint of the limb. 
  Notice that the torque on the platform due to the propellers is smaller for the sagittal case (center), but the platform can tilt much quicker (top) thanks to the dynamic tail assistance (bottom).}
  \label{fig:exp_flying_displacement}
\end{figure}

We show in \figref{fig:exp_flying_displacement} a comparison of the sagittal and coronal trajectories with the same time interval of $2\si{\second}$.
In the sagittal case, the platform can tilt much quicker (top plot) while applying a smaller torque on the platform due to the propellers (center plot). 
We attribute this difference to the dynamic tail assistance, as seen in the bottom plot where the tail contributes over $2\si{\newton\meter}$ of torque.
It might be worth remarking that this assistance maneuver has been discovered and computed by the \gls{ocp}; in other words, it has not been enforced or suggested by any other means than optimality.
The dynamic and static assistance can also be appreciated in \figref{fig:sequence-displacements}. 
We observe rapid and ample tail movements in the sagittal case, typical of dynamic assistance.
We also observe static tail assistance in the coronal case, again discovered by the \gls{ocp}, where the tail configuration changes slowly to achieve a  better centered and slightly higher  \gls{com}, thereby easing the task of tilting the robot.
This tilting needs to be performed exclusively via the propellers' thrust differential.

\begin{figure}[t]
  \centering
  \includegraphics{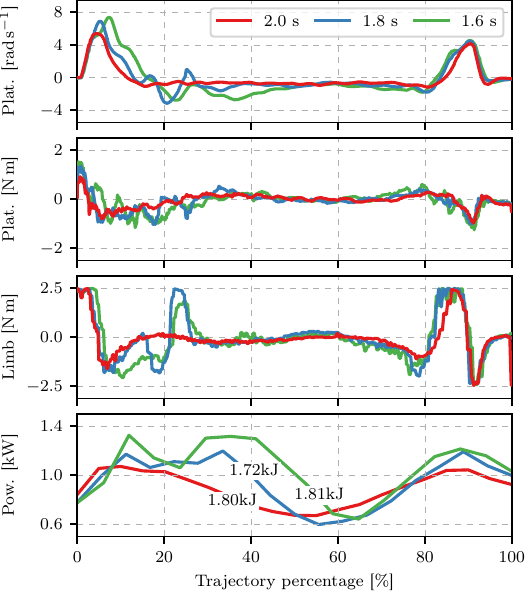}
  \caption{Borinot's performance for the increasingly aggressive sagittal trajectory. 
  Notice that the horizontal axis is normalized for the three trajectory durations. 
  From top to bottom. 
  \emph{First:} platform's angular speed in the sagittal plane. 
  \emph{Second:} torque applied to the platform only due to propeller thrusts. 
  \emph{Third:} torque of the first joint in the limb. 
  The slow 2s trajectory exhibits a smooth action of the limb, representing a graceful execution (see \secref{sec:degrees-agility}). 
  The 1.8\,s, and 1.6\,s trajectories progressively saturate the limb's torques (at $\pm2.5$\,Nm), representing more aggressive executions.
  \emph{Fourth:} Power consumption and total electro-mechanical work (\ie~energy). 
  The graceful trajectory (red) exhibits a smoother power curve, although the total energy consumption is equivalent for the three cases.
  }
  \label{fig:exp_flying_x_displacement}
\end{figure}

We conducted a second set of experiments to investigate the extent to which the limb's contribution can enhance the platform's performance during increasingly aggressive trajectories in the sagittal plane.
We now ask Borinot to go to and come back from the waypoint at increasing speeds, that is, with decreasing time intervals of $2.0\,\si{\second}$, $1.8\,\si{\second}$, and $1.6\,\si{\second}$.
We report the results in \figref{fig:exp_flying_x_displacement}.

We find that for the $2.0\,\si{\second}$ trajectory, the limb's action is already saturated at $2.5\,\mathrm{Nm}$ for the initial 5\% of the time.
During this initial kick, and since the tail torque is already saturated, the additional angular speed of the platform observed for the $1.8\si{\second}$ and $1.6\si{\second}$ trajectories is  due to the differential action of the propellers (second plot).
After this and up to 30\% of the time, we observe an increasing contribution of the tail as the time of the experiment shortens.
This tail assistance controls the tilting angle, which reaches values close to 60$^\circ$ for the 1.6\,s trajectory.
We also observe that the limb's saturation persists for extended periods during these quicker trajectories.
These results suggest that the trajectory of $2.0\si{\second}$ marks the threshold between a graceful motion (see \secref{sec:degrees-agility}), where controls remain smooth, and an aggressive motion, where the controls become saturated for extended intervals, and the transitions are sharper.
This increasing aggressiveness can be better appreciated in the accompanying video.

A final commentary concerns power and aggressiveness. 
More aggressive motions require more power, as observed in \figref{fig:exp_flying_x_displacement}-bottom. 
Interestingly, such an increase in power is well compensated by a shorter execution time, leading to very similar energy consumption for completing the three tasks.

\subsection{Flying manipulation: limb as an arm}
\label{subsec:flying_manipulation}

\begin{figure}[t]
  \centering
  \includegraphics[width=0.75\columnwidth]{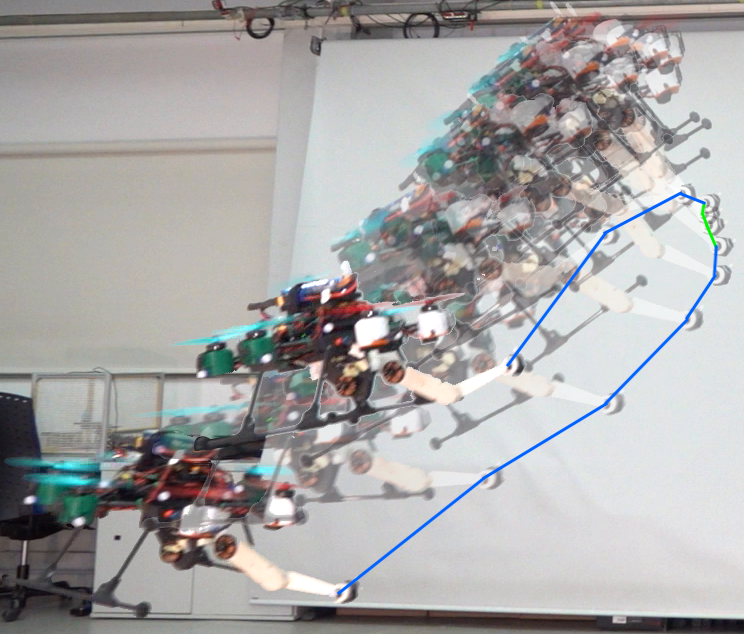}\\
  \vspace{1ex}
  \includegraphics{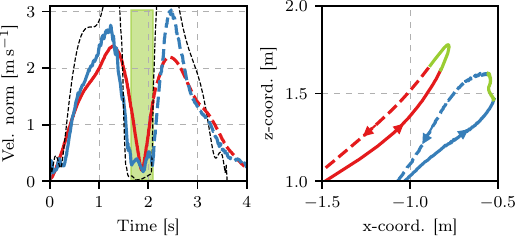}
  \caption{Borinot's performance for the dynamic manipulation task. 
  \emph{Top:} sequence of configurations, with overimposed \gls{ee} trajectory showing approach (lower blue), task (green), and departure (upper blue) sections. 
  \emph{Bottom left:} speed (velocity norm) of the platform (red), EE (blue), and the EE's reference from the \gls{ocp} (dashed black). 
  The EE drastically reduces speed during 500\,ms (flat of the blue curve in green area) while the platform performs an unstable maneuver at higher speeds governed by gravity. 
  \emph{Bottom right:} detail of the sagittal trajectories of the base (red) and EE (blue) around and during the EE positioning task (green). 
  Notice that the slightly deficient precision of the EE position is due to the lack of a closed-loop control at the task level (see \secref{sec:control_architecture} and \figref{fig:control_architecture}), but that the dynamics of the task are properly captured.
  }
  \label{fig:exp_flying_agile_ee}
\end{figure}

\glsreset{ee}
In this experiment, we demonstrate an agile approach for a manipulation task.
The task is to maintain the \gls{ee} at a fixed position for a brief but non-trivial period  ($500\,\si{\milli\second}$) while the platform is prevented from hovering, which we achieve by demanding a pitch angle of 45\,$^\circ$ during the task.
This way, we emulate the conditions for a manipulation in which the robot base cannot stop and hover, such as when having to pick an object from a wall with an arm that is shorter than the propellers.
This forces a dynamic maneuver.

The results are shown in \figref{fig:exp_flying_agile_ee}. 
We observe that it is possible to accomplish this manipulation task by executing an agile maneuver where the tilted platform follows a parabolic trajectory governed by gravity. 
The manipulation occurs around the instant when the platform reaches the apex of this trajectory. 
During this short time, the arm continuously adjusts to ensure that the \gls{ee} remains at the desired position.

\figref{fig:exp_flying_agile_ee} indicates that the maneuver dynamics are properly captured. 
That is, the platform does a parabolic trajectory while the \gls{ee}'s position remains within a short range from a fixed point, and its velocity remains very low.
However, to maintain the \gls{ee} in a truly fixed position, its velocity should be zero.
This is not the case, and we attribute it to the lack of closed-loop control at the task level
(see \figref{fig:control_architecture}), which prevents the positioning task from reaching the desired accuracy.
Nonetheless, the results indicate that the approach is feasible and effective for an agile manipulation task. 
We discuss this control issue further in the conclusions.

\subsection{Hybrid aerial-contact locomotion: limb as a leg}
\label{subsec:exp_aerial_contact_locomotion}

\begin{figure*}[t]
  \centering
  \includegraphics[width=0.135\textwidth]{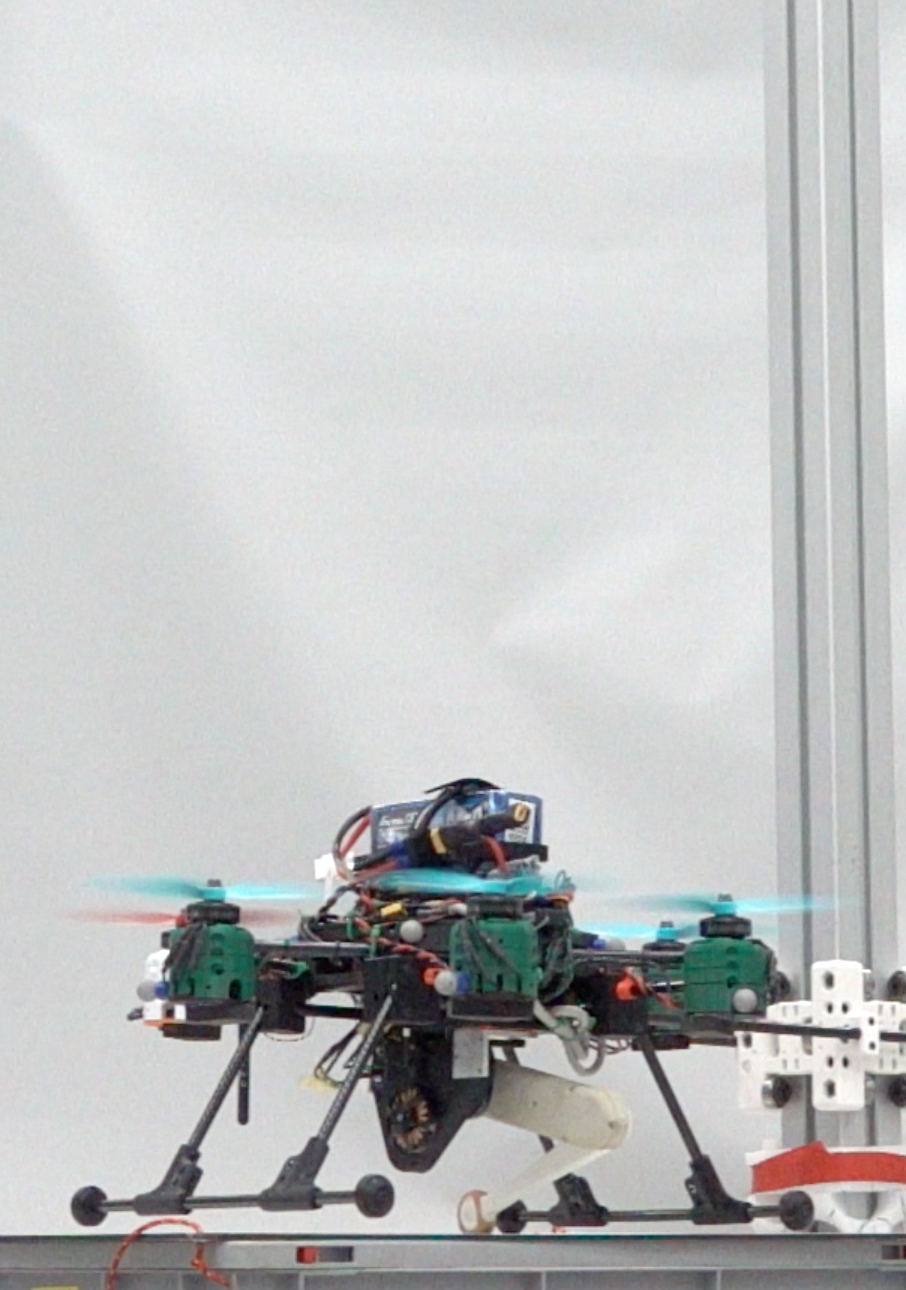}
  \includegraphics[width=0.135\textwidth]{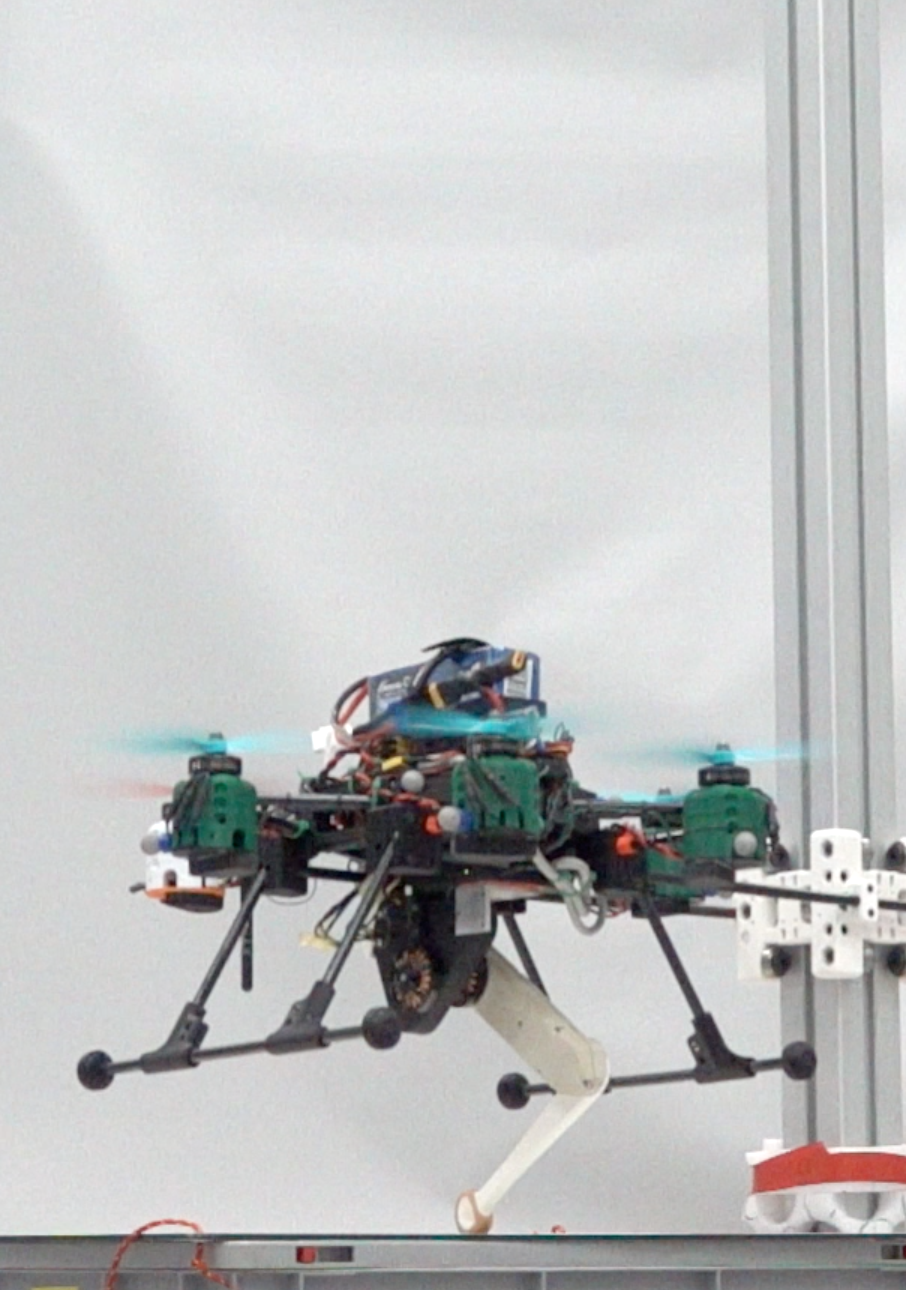}
  \includegraphics[width=0.135\textwidth]{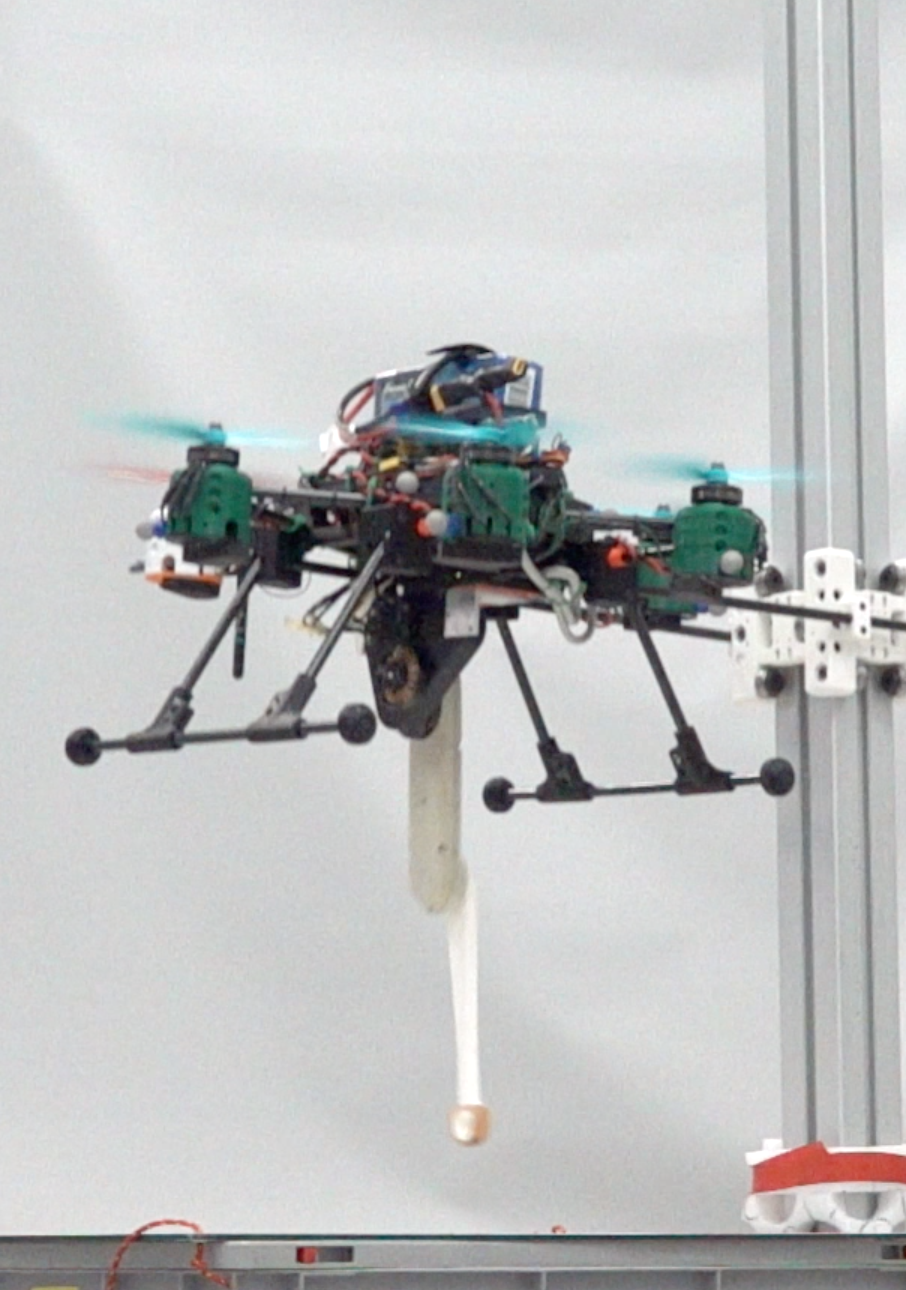}
  \includegraphics[width=0.135\textwidth]{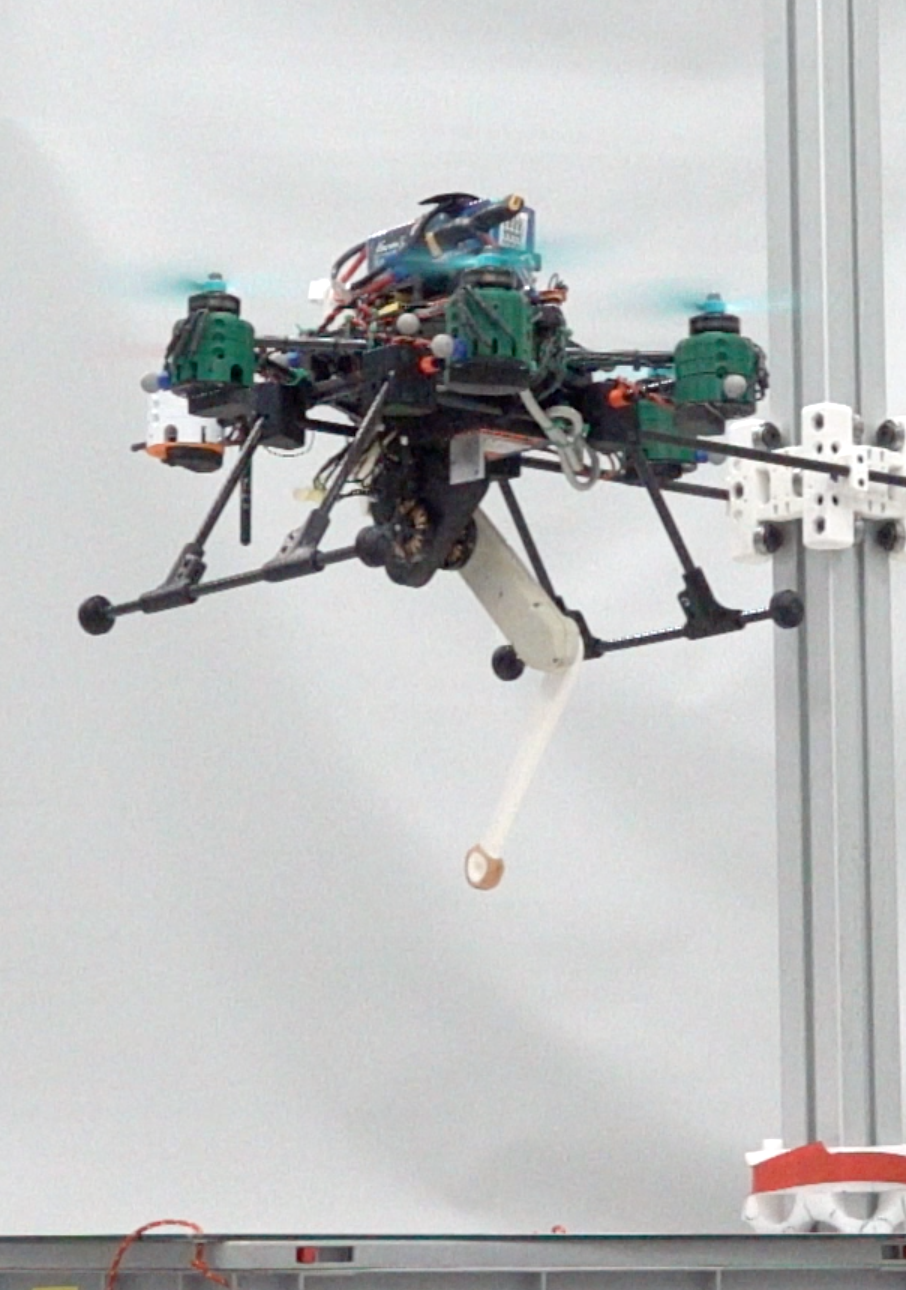}
  \includegraphics[width=0.135\textwidth]{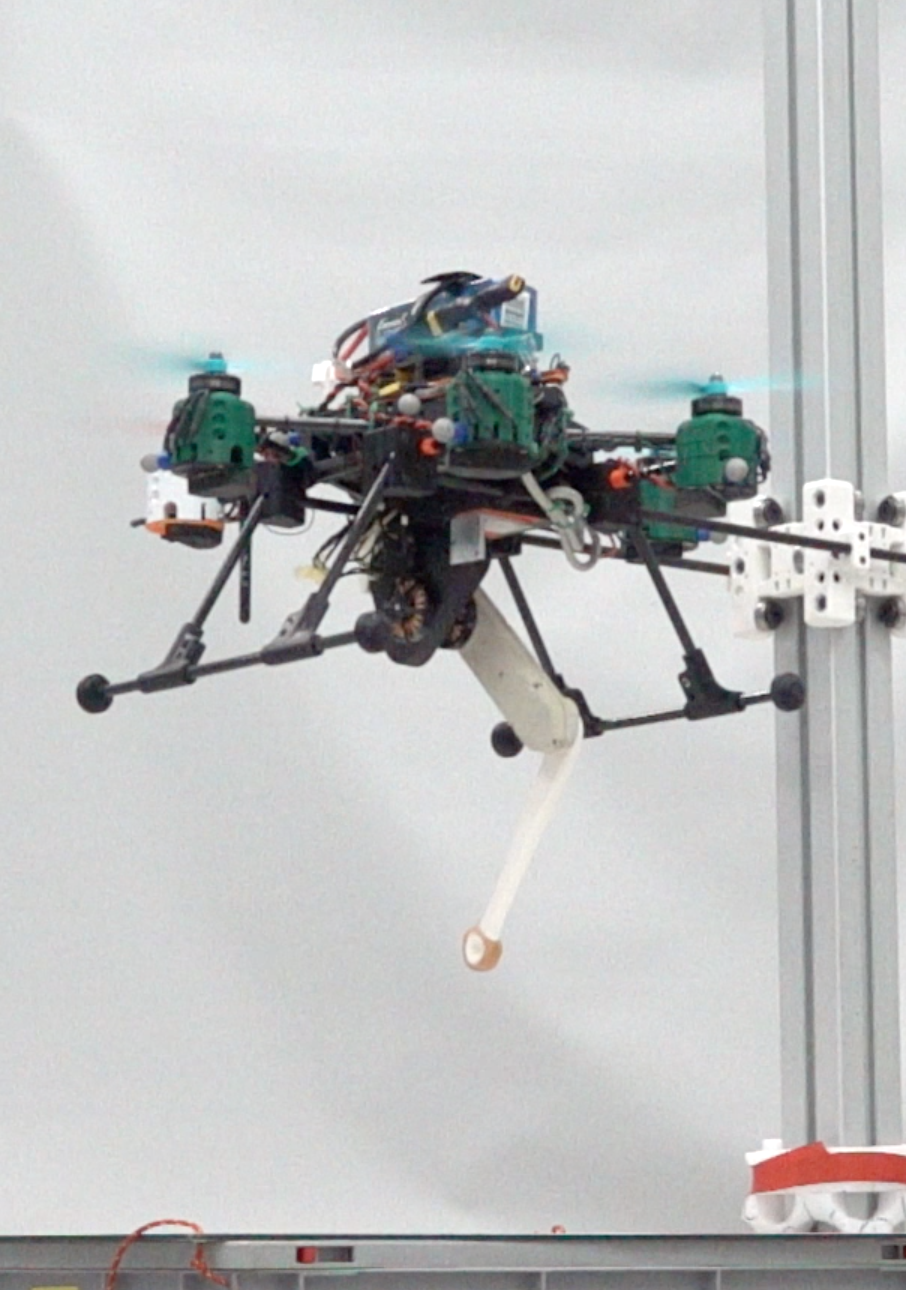}
  \includegraphics[width=0.135\textwidth]{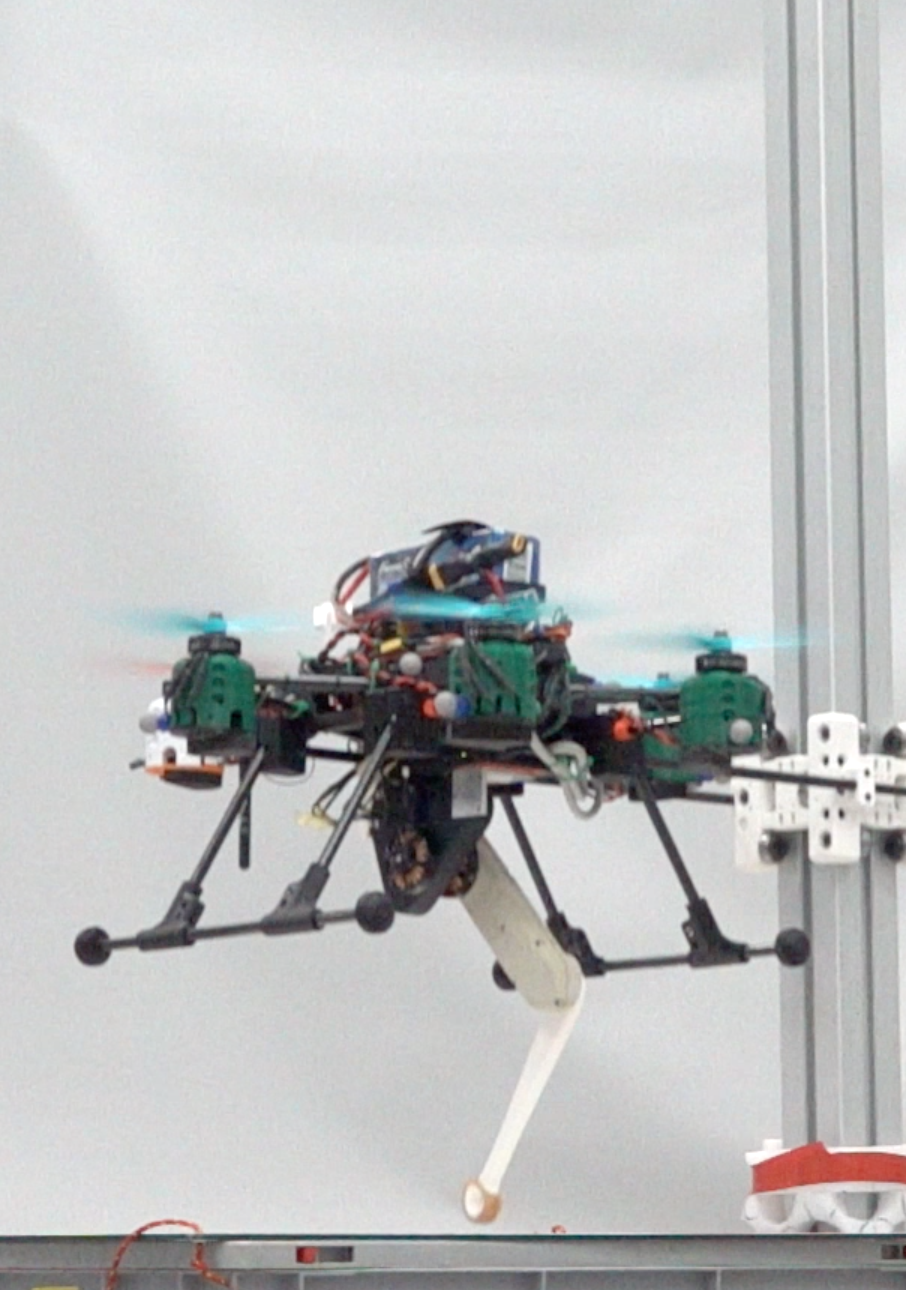}  
  \includegraphics[width=0.135\textwidth]{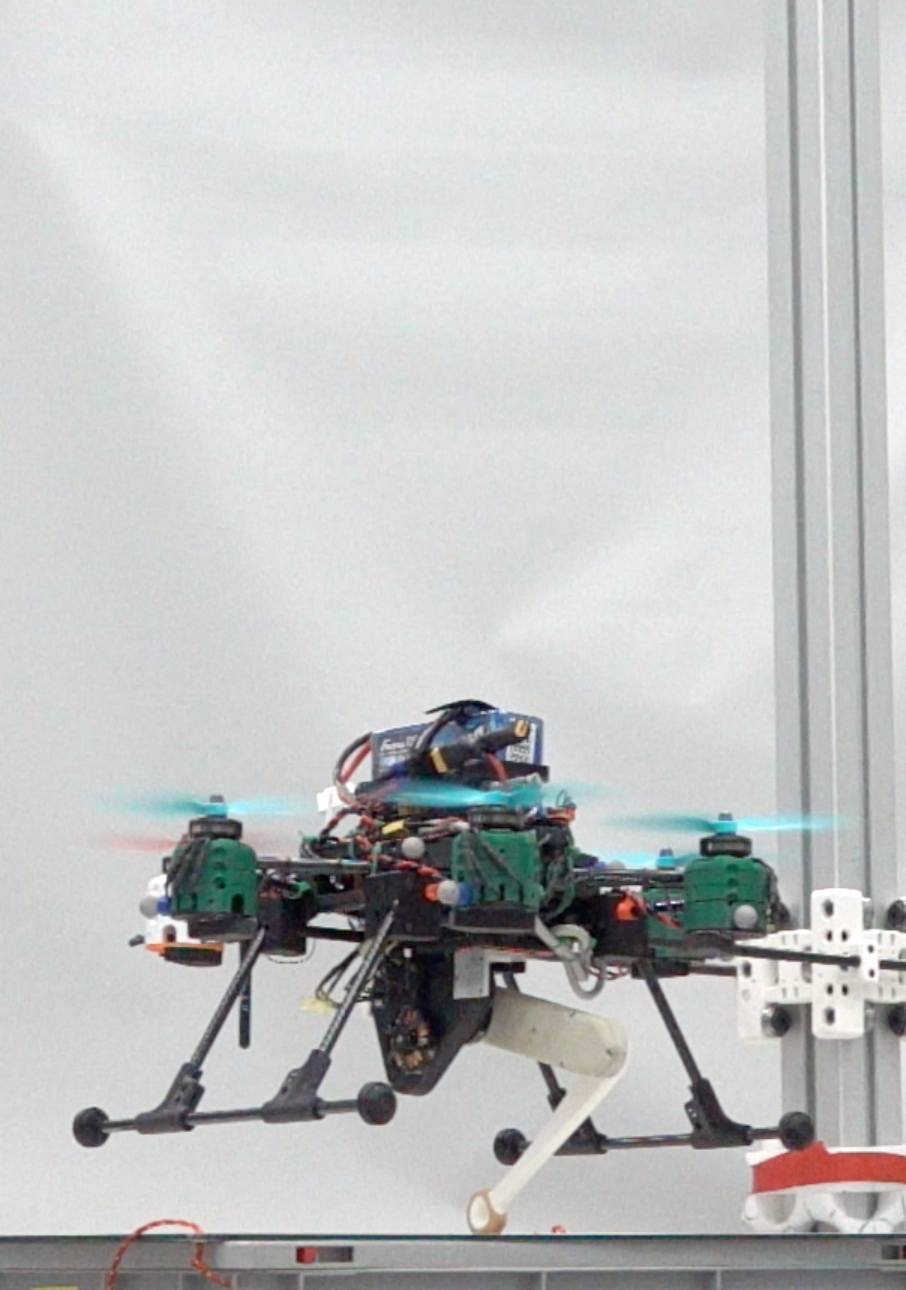}
  \caption{Jump-and-fly sequence. 
  The thrust is set somewhat below the robot's weight, and the leg is driven at full torque, producing a jump (frames 1 and 2). 
  Once in flight (frames 3 to 5), the leg is set to a mid-impedance compliant mode to allow a soft landing (frames 6 and 7).}
  \label{fig:exp_jump_trajectory}
\end{figure*}

\begin{figure}[t]
  \centering
  \includegraphics{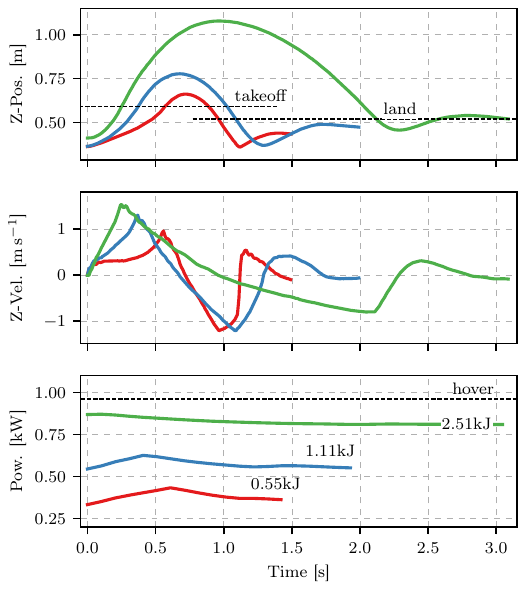}
  \caption{Three jump-and-fly trajectories at thrust levels of 50\% (red), 70\% (blue) and 90\% (green) of the total weight. 
  \textit{Top:} z-positions, showing the takeoff and landing heights (which differ due to the different leg configurations). 
  \textit{Center:} z-velocities showing slower-than-gravity accelerations. 
  The initial height offset of the 90\% curve and its slowed decay are due to the ground effect (increased lift close to the ground).
  \textit{Bottom:} Power consumption and total energy for one jump-and-fly step, also showing the power for a hovering flight (\ie, 100\% thrust-to-weight level). 
  The jump-and-fly strategy is much more efficient than pure flight and might, for this reason, be the locomotion mode of choice in some scenarios.}
  \label{fig:exp_jump_results}
\end{figure}

The hybrid aerial-contact locomotion experiment consists of Borinot jumping using its limb as a leg while having its propellers holding a fraction of its weight.
We have attached the robot to a carriage that can move through a vertical rail (\figref{fig:exp_jump_trajectory}).
This allows us to eliminate the necessity of robot control and concentrate on showing the electro-mechanical capacity of Borinot for the jump-and-fly locomotion mode.

The jump-and-fly sequence of moves can be observed in \figref{fig:exp_jump_trajectory}.
To perform the jump, we set the platform's motors at a constant thrust below the total weight (Borinot + carriage).
Then, starting with the folded leg, we apply full torque to its joints ($2.7\si{\newton\meter}$) until the leg is fully stretched and the robot goes airborne.
During the flight, we bring the leg to a semi-folded configuration with medium impedance to allow it to touch the ground in a compliant soft landing.

We show in  \figref{fig:exp_jump_results} the results for three different jumps with the thrust set respectively at $50\%$, $70\%$, and $90\%$ of the total moving weight.
As expected, the robot is able to jump higher for higher thrust values and for a longer time (top plot).
During the airborne phase, the three trajectories should have a perfect parabolic shape, with a linear decay of velocity corresponding respectively to 50\%, 30\%, and 10\% of gravity (center plot).
This behavior can be distorted by the apparition of the ground effect\footnote{\url{https://en.wikipedia.org/wiki/Ground_effect_(aerodynamics)\#Rotorcraft}}, which is especially evident in the case of the $90\%$ thrust jump.

The bottom plot of \figref{fig:exp_jump_results} shows the total power required for each jump, together with a reference of the power required for hovering, which is 927W if we account for the carrier's weight (around 700\,g).
We see that the locomotion power required increases as we increase the contribution of the flight with respect to the jump. 
In other words, jump-and-fly locomotion is much more efficient than pure flight, and the more we can rely on the legs, the better. 
This reveals that exploring such hybrid locomotion modes is beneficial, for flying is an expensive endeavor for creatures with poor aerodynamics, but walking, running, or jumping alone might be insufficient to achieve certain tasks.
The fact that some animals perform this kind of hybrid locomotion suggests that for specific body architectures and in certain situations (\eg~escaping from predators or climbing steep slopes), this is the preferred mode of locomotion \cite{wei-19-fly-jump,Vidyasagar_2015,birn2014don,tobalske2007aerodynamics,badri-2022-birdbot}.
This should also apply to robots.

\section{Discussion}
\label{sec:discussion}




In this paper, we have introduced Borinot, an open-source robot designed for research on agile aerial-contact loco-manipulation.
Throughout our study, we successfully validated the electromechanical capabilities of this robot by conducting a series of experiments to test the different operation modes.
Based on this work, we should direct our future efforts at two levels. 
The first level comprises engineering improvements at both hardware and software levels. 
The second level involves research in the field of motion control to fill the gaps that remain uncovered by this work.
Both are developed in the following paragraphs.

In terms of the design of the robot, we believe that the improvements should lead to a lighter and cheaper robot with fewer custom-made software repositories to maintain.
To achieve a lighter robot, we should target the platform's 3D-printed parts, which currently contribute to $25\%$ of the total weight.
In line with the weight reduction goal, we aim to evaluate alternative onboard computers that are lighter and more cost-effective compared to the current solution, which is half the cost of the total platform.
Furthermore, to eliminate the need for maintaining our custom version of \textsc{px4}, we intend to utilize the existing firmware version that already contains the capabilities we have developed in our version. 
This will streamline maintenance and align our system with established off-the-shelf solutions.
Still related to the flight controller, we believe exploring the use of more affordable hardware is worthwhile, as suggested in \cite{foehn_AgiliciousOpensourceOpenhardware_2022}.

Regarding motion control, we have presented a basic control architecture based on \gls{mpc}.
This allowed us to validate the robot, especially for agile flying locomotion, but left several topics unaddressed, which individually require intensive research.
First, the control architecture does not consider contacts.
The theoretical tools for including them in the \glspl{ocp} are known and used in legged locomotion. 
Fundamental difficulties arise because it is difficult or impossible to predict the time of contact. 
This can be a very acute problem under high dynamics, requiring accurate contact surface reconstruction and/or fast and reliable contact detection and fast control reaction.
Second, the control architecture lacks precision for manipulation tasks.
This should improve by endowing the controller loop with the notion of task, recovering at the same time the navigation phases.
This would allow the \gls{mpc} to recompute maneuvers to accommodate the tasks with precision.
Here, we face difficulties related to the discontinuities arising when the tasks first appear in the \gls{mpc} horizon.
In such cases, the new \gls{ocp} happens to be very different from the last, and the solver requires a sudden increase in the number of iterations, sometimes even failing to converge.
These issues can sometimes be tackled with heuristics or fine-tuning. 
Other times they require redesigning some components, a deeper theoretical insight, and/or adopting radically different techniques such as reinforcement learning in certain parts of the control scheme.

To conclude, our ultimate interest is to demonstrate hybrid motion modes that are particular to aerial-contact loco-manipulation.
With the help of the research platform Borinot, we believe we have established a consistent foundation to pursue this research.
By making it open source, we invite any interested actor to join the effort.






  

%
  

\bibliography{files/references}

\end{document}